%% file: main.tex
\RequirePackage{iftex}
\ifXeTeX
  \providecommand{\pdfmapline}[1]{}
\fi
\documentclass{fairmeta}
\ifPDFTeX
  
\fi
\ifXeTeX
  \RequirePackage{fontspec}
\fi

\ifdefined\pdfinfoomitdate\pdfinfoomitdate=1\fi
\ifdefined\pdfsuppressptexinfo\pdfsuppressptexinfo=15\fi
\ifdefined\pdftrailerid\pdftrailerid{}\fi

\usepackage{amsmath,amssymb,mathtools}
\usepackage{amsthm}
\usepackage{tabularx,array,longtable}
\usepackage{enumitem}
\usepackage{float}
\usepackage{flafter}
\usetikzlibrary{decorations.pathreplacing}
\usepackage{url}

\newcommand{\R}{\mathbb{R}}
\newcommand{\TV}{\operatorname{TV}}
\newcommand{\ind}{\mathbf 1}
\newcolumntype{Y}{>{\raggedright\arraybackslash}X}
\setlist{leftmargin=1.25em,itemsep=0.12em,topsep=0.2em}
\renewcommand{\arraystretch}{1.03}
\theoremstyle{definition}
  \newtheorem{definition}{Definition}
  \theoremstyle{plain}
  \newtheorem{lemma}{Lemma}
  \newtheorem{theorem}{Theorem}
  \newtheorem{corollary}{Corollary}
  \newtheorem{proposition}{Proposition}
  \theoremstyle{remark}
  \newtheorem{remark}{Remark}

\title{Beyond Update Counts: A Control Map for Evaluating Test-Time Model Updates}
\author[*]{Zehao Zhao}
  \author{Shangjian Yin}
  \author{Rui Liu}
  \author{Chongyang Bai}
  \author{Mark Ibrahim}
  \author{Kavosh Asadi}
  \author{Yuchen Lu}
  \author{Hang Cui}
  \author{Hamed Firooz}

  \affiliation{Meta AI}
  \contribution[*]{Work done at Meta.}
\hypersetup{
  pdftitle={Beyond Update Counts: A Control Map for Evaluating Test-Time Model Updates},
  pdfauthor={Zehao Zhao, Shangjian Yin, Rui Liu, Chongyang Bai, Mark Ibrahim, Kavosh Asadi, Yuchen Lu, Hang Cui, Hamed Firooz}
}
\abstract{\input{sections/abstract.tex}}
\date{Oct 23, 2026}

\begin{document}
\raggedbottom
\setlength{\emergencystretch}{3em}
\maketitle

\input{sections/01_introduction.tex}

\input{sections/02_control_map_formulation.tex}

\input{sections/03_update_reachability.tex}

\input{sections/04_behavioral_effects.tex}

\input{sections/05_memory_feasibility.tex}

\input{sections/06_cross_setting_generalization.tex}

\input{sections/07_conclusion.tex}

\input{sections/backmatter/statements.tex}

\bibliography{bibliography/references,bibliography/additional_references}
\bibliographystyle{assets/plainnat}

\clearpage
\beginappendix
\input{sections/appendices/audit_and_empirical_details.tex}
\input{sections/appendices/formal_proofs.tex}

\end{document}

%% file: sections/abstract.tex
  Test-time updating lets a language model change persistent state after
  training---such as model weights, adapter parameters, or a memory
  module---before or while answering queries. Existing evaluations often
  compare systems using update count or frequency and performance measured
  immediately afterward. These summaries omit what is updated, where, when, and in
  what order. The same total can therefore hide an important difference: one
  system may answer after one update while another answers after two. We introduce the \emph{Control Map}, an evaluation framework that records the committed updates preceding each scored answer and separately evaluates behavioral changes,
  immediate learning (\emph{acquisition}), later recall (\emph{retention}), preservation of unrelated behavior (\emph{locality}), and performance in new settings. We prove when counts alone determine final state and show, under stated assumptions, how differences between pre-answer histories limit differences in average scores. Using \emph{ControlMap}, our experiments reveal three insights. First, update scheduling matters: at a 760M-parameter checkpoint, 24 schedules with the same update count yield 19.7\%--34.2\% accuracy. Second, learning an edit does not guarantee that it lasts or remains contained: all six selected CounterFact LoRA settings learn the new information, while ROME retains edits at all 13 operating points but fails locality at each. Third, no editor performs best consistently across models, benchmarks, and tests. Evaluations should therefore report update histories and assess acquisition, retention, and locality separately.


%% file: sections/01_introduction.tex
\section{Introduction}
\label{sec:introduction}
Test-time model updates let a model change persistent state after
  training---such as model weights, adapter parameters, or a memory
  module---before or while answering queries. This includes systems based on
  fast weights, test-time training, adaptive memory, and model editing
  \citep{schlag2021linear,sun2025tttlayers,behrouz2025titans,meng2022rome}.
  A policy, or update plan, must choose what content to write, which parameters
  or layers to modify, which update mechanism to use (called the \emph{writer}
  below), how strongly and often to update, and when those updates occur relative
  to each query. Representative studies vary or report layer location, update
  timing or frequency, update strength, and the number of edits processed
  together
  \citep{tandon2025endtoend,lee2023surgical,wakayama2026decision,
  meng2023memit,yoon2024batch}. Yet comparisons based only on policy labels,
  update frequencies, or total update counts compress these choices into a small
  number of summary values.

\begin{figure}[H]
  \centering
  \scalebox{1.10}[1]{%
    \includegraphics[
      width=0.91\linewidth,
      trim=10 90 10 110,
      clip
    ]{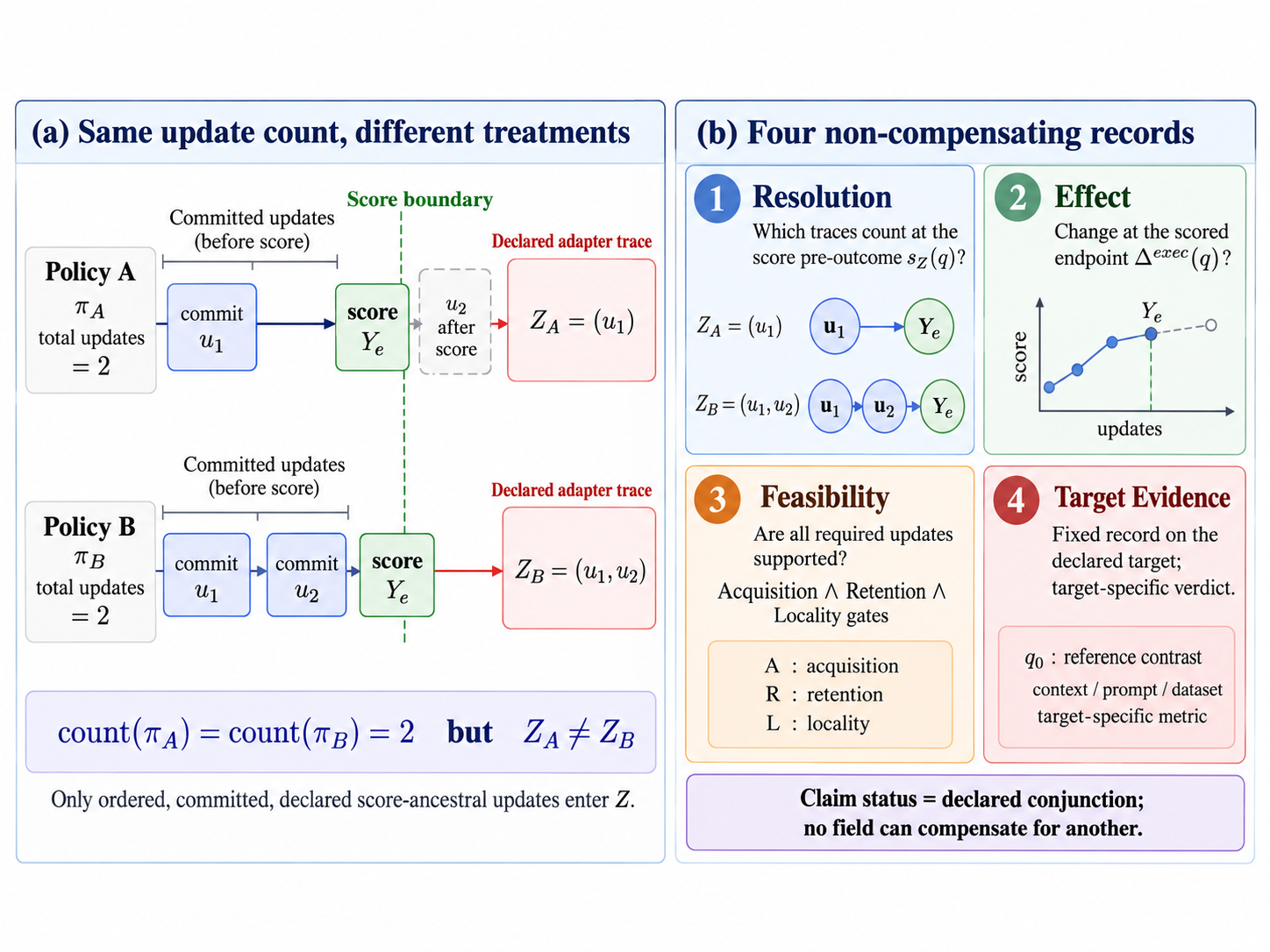}
         }
    \caption{\textbf{The same update count can yield different histories
    at an answer.} Both plans perform $u_1$ and $u_2$, but only plan B performs
    $u_2$ before scoring, so their recorded histories are $(u_1)$ and
    $(u_1,u_2)$ under the stated tracing rule. The Control Map reports update
    history, measured effect, memory quality, and results in a new setting
    separately. The formal record also stores cost; values are schematic.}

    \label{fig:controller-overview}
  \end{figure}

Two systems can perform the same total number of updates over a full run, yet
  an answer may be produced after one update in one system and after two in the
  other (Figure~\ref{fig:controller-overview}). This difference can be
  consequential: if the second update contains information needed for the query,
  one system is evaluated before it can use that information, whereas the other
  is evaluated afterward. Describing both as ``two-update'' systems therefore
  hides a real difference in the information available when they answer.
  Conversely, different update plans can produce the same ordered update history
  before an answer. In that case, comparing their policy labels cannot reveal
  whether the intended difference between the plans matters.

  Establishing that the histories differ is only the first step. A complete
  evaluation must ask four questions: (1) which ordered updates were completed
  before each answer, and did the compared plans differ; (2) did those differences change the model's answer or evaluation score; (3) was new information learned immediately
  (\emph{acquisition}), recalled later (\emph{retention}), and learned without
  changing unrelated behavior (\emph{locality}); and (4) does the conclusion hold
  in a new model or benchmark setting? Evidence for one question cannot answer
  the others: different histories do not guarantee a behavioral change,
  immediate learning does not guarantee later recall or locality, and a result in
  one setting does not guarantee the same result in another.

We address these questions with the \emph{Control Map}, an evaluation framework
  rather than a new update algorithm. Before examining outcome scores, the
  evaluator specifies a tracing rule that identifies which completed updates may
  affect each answer. For every update plan and example, the framework records
  these updates in order and groups plans that produce the same history. Its
  conclusions therefore depend on the stated tracing rule; the framework does not
  claim to discover every possible path to an answer automatically. It reports
  separately whether the histories differ, whether behavior changes, whether
  acquisition, retention, and locality each meet their requirements, and whether
  the conclusion is supported by direct evaluation in a new setting. This framework brings together three lines of prior work: methods that update
  model state during inference, evaluations of model editing and memory, and
  methods for identifying which intervention or computation actually affects an
  output.

\subsection{Related work and positioning}
\label{sec:related-work}

 \paragraph{Stateful test-time adaptation.}
  Fast weights, test-time-training layers, and hierarchical memory systems study
  how models should update state during inference
  \citep{schlag2021linear,yang2024deltanet,sun2025tttlayers,
  behrouz2025titans,behrouz2025nested,behrouz2025miras}. Related work treats
  update frequency, layer location, routing, and update strength as design
  variables
  \citep{tandon2025endtoend,hu2025testtimelearning,sim2026ponder,
  wakayama2026decision}. These studies primarily develop mechanisms for updating
  state. We study the complementary evaluation question: once an update method
  and plan have been chosen, which updates were actually completed before each
  evaluated answer?

\paragraph{Model editing and memory evaluation.}
  ROME, MEMIT, AlphaEdit, and model-editing toolkits evaluate edit
  success, generalization to related prompts, locality, and propagation to
  related facts
  \citep{meng2022rome,meng2023memit,fang2024alphaedit,wang2023easyedit,
  zhang2024knowedit,cohen2024rippleedits}. Recent work further shows that
  reported editor rankings can change with the evaluation metric, whether edits
  are processed individually or in batches, and other protocol choices
  \citep{gupta2024unified,pohl2025principled,yoon2024batch,
  liu2026editlocality}. UniEdit extends evaluation to open-domain, multi-step
  test questions \citep{chen2025uniedit}. This literature establishes the need
  to evaluate several dimensions of model editing. We add a complementary
  execution question: which ordered updates were completed before each evaluated
  answer? The Control Map links each result to that history and treats immediate
  learning, later recall, and preservation of unrelated behavior as separate
  requirements.

 \paragraph{Assignments and computational provenance.}
  Causal-inference work distinguishes an assigned intervention from what a case
  actually experiences
  \citep{vanderweele2013versions,aronow2017generalinterference,
  savje2023misspecified,adhikari2025learningexposure,huber2026learningtesting},
  while program-analysis work traces which earlier operations can influence an
  output
  \citep{weiser1981slicing,korel1988dynamic,zhang2020dynamicslicing,
  lee2021causalprogram,tian2026aiprovenance}. These literatures explain why an
  assigned label may not describe what actually happened and how relevant
  operations can be traced. However, neither provides an evaluation
  that connects each model answer to the ordered updates completed before it and
  then separately tests behavioral change, retention, locality, and performance
  in a new setting. The Control Map provides this connection using a tracing rule
  fixed before scores are examined. Its guarantees depend on that rule capturing
  every way the update plan can affect the score, and claims about a new setting
  require direct evaluation there rather than assuming that the original result
  will carry over \citep{pearl2014external}.
  
 \paragraph{Contributions.}
  Motivated by these complementary lines of work and the gaps identified above,
  we make three contributions:

  \begin{enumerate}[leftmargin=1.5em]
    \item \textbf{Evaluation framework.}
      We introduce the Control Map, which records the ordered updates completed
      before each answer and separately reports whether plans produce different
      histories, change behavior, satisfy acquisition, retention, and locality
      requirements, and continue to work in a new setting.

      \item \textbf{Formal guarantees.}
      We show that
      greater overlap between recorded histories places a tighter bound on
      average-score differences under stated assumptions. We also characterize
      when update counts determine final state and extend the score bound to a
      known amount of mismatch in the scoring process.

      \item \textbf{Empirical evidence.}
      We find that layer placement, query timing, and the number of updates affect
      outcomes in setting-dependent ways. At one checkpoint, 24 equal-count
      schedules span 19.7\%--34.2\% accuracy, while matching the same updates and
      order produces 1,600 exact terminal-state and locality ties. Across six selected CounterFact settings, acquisition passes in all six, but retention and locality pass in none: information learned immediately can later be forgotten or alter unrelated answers. Method rankings can also change across models, benchmarks, and tests.
\end{enumerate}

Section~\ref{sec:framework} formally defines update plans, pre-answer histories,
  and the Control Map. Section~\ref{sec:execution} asks when different plans
  actually produce distinguishable histories. Section~\ref{sec:effects} studies
  the effects of layer placement, timing relative to the query, and the number of
  updates completed before a query. Section~\ref{sec:feasibility} separates
  immediate learning from later recall and preservation of unrelated behavior.
  Section~\ref{sec:transport} tests whether conclusions continue to hold across
  new models and benchmarks. Section~\ref{sec:conclusion} concludes. Additional
  experimental details, threshold definitions, sensitivity analyses, and
  provenance records appear in
  Appendices~\ref{app:audit-contract}--\ref{app:artifact}; complete proofs appear
  in Appendix~\ref{app:formal-proofs}.

%% file: sections/02_control_map_formulation.tex
\section{The Control Map and Problem Formulation}
\label{sec:framework}

\subsection{Stateful policies and operating points}

A setting is $c=(M,\mathcal T,e,\mathcal B)$, where $M$ is the model,
$\mathcal T$ the task and interface, $e$ the output being evaluated, and
$\mathcal B$ an optional cost budget; a row $r$ is one evaluation example. An
operating point pairs a setting with a fully specified update policy and
configuration.

 \begin{definition}[Policy and declared pre-answer update history]
  \label{def:policy-history}
  Fix a setting $c=(M,\mathcal T,e,\mathcal B)$ and a filtration
  $(\mathcal F_t)_t$ containing only information available before update
  opportunity $t$. A test-time update policy $\pi$ consists of the action kernel
  \begin{equation}
  a_t=(g_t,x_t,\ell_t,w_t,\eta_t)
  \sim\pi_{\rm act}(\cdot\mid\mathcal F_t,c;\kappa,\phi),
  \label{eq:control-action-v94}
  \end{equation}
  together with its cadence $\kappa$, query-relative phase $\phi$, and commit
  rule. Here $g_t$ admits an update, $x_t$ selects available content, $\ell_t$
  selects parameter support, $w_t$ selects the writer, and $\eta_t$ its
  strength. With commit indicator $\delta_t$, model state $\theta_t$, and writer
  state $\omega_t$, the complete mutable state obeys
  \begin{equation}
  (\theta_{t+1},\omega_{t+1})=
  \begin{cases}
  F_{w_t,\ell_t,\eta_t}(\theta_t,\omega_t;x_t),&\delta_tg_t=1,\\
  (\theta_t,\omega_t),&\delta_tg_t=0.
  \end{cases}
  \label{eq:state-update-v94}
  \end{equation}
  For row $r$ and scored endpoint $e$, let $\Lambda_{r,\pi,e}$ be the ordered log
  of committed transitions preceding $e$. A dependency specification
  $\mathcal D_{c,e}$ maps
  this log to the canonical ordered sequence
  \[
  z_{r,\pi,e}:=\mathcal D_{c,e}(\Lambda_{r,\pi,e})
  \]
  of committed transitions it declares able to influence the score. We call
  $z_{r,\pi,e}$ the \emph{declared pre-answer update history}. Policies $\pi$
  and $\pi'$ are history-equivalent on $(r,e)$, written
  $\pi\sim_{r,e}\pi'$, iff $z_{r,\pi,e}=z_{r,\pi',e}$.
  \end{definition}

\subsection{Pre-answer histories and comparison resolution}

 \begin{definition}[Contrast-specific history resolution and score sufficiency]
  \label{def:trace-resolution}
  Fix the scored endpoint $e$ and the dependency specification of
  Definition~\ref{def:policy-history}. Let $\Pi$ be a finite policy catalogue and
  $\mathcal R$ a finite row set with
  policy-independent weights $w_r\geq0$ and $\sum_rw_r=1$. Assume the history
  and outcome spaces are finite or standard Borel. For each
  $\pi\in\Pi$, let $K_{\pi,e}(\mathrm dz\mid r)$ be the conditional law of its
  declared history. Let $R$ denote a row sampled according to the weights and
  $Z$ its declared history, and define
  \[
  P_{R,Z}^{\pi}(\{r\}\times A)=w_rK_{\pi,e}(A\mid r).
  \]
  A contrast $q$ is a weighted comparison of policies whose coefficients sum
  to zero. For nonzero $q\in\R^{|\Pi|}$ with $\mathbf1^\top q=0$, write
  $q^+_\pi=\max\{q_\pi,0\}$ and $q^-_\pi=\max\{-q_\pi,0\}$. Since
  $\lVert q^+\rVert_1=\lVert q^-\rVert_1=\lVert q\rVert_1/2$, define the two
  contrast-induced probability laws by
  \[
  P_{R,Z}^{q^\pm}=\sum_{\pi\in\Pi}
  \frac{2q_\pi^\pm}{\lVert q\rVert_1}P_{R,Z}^{\pi}.
  \]
  For probability measures $P,Q$, use the convention
  \[
  \TV(P,Q):=\sup_A|P(A)-Q(A)|
  =\frac12\int\left|\frac{\mathrm dP}{\mathrm d\lambda}
  -\frac{\mathrm dQ}{\mathrm d\lambda}\right|\mathrm d\lambda,
  \]
  where $\lambda=P+Q$. The
  \emph{history resolution} of $q$ is
  \[
  s_Z(q):=\TV(P_{R,Z}^{q^+},P_{R,Z}^{q^-}).
  \]
  For deterministic histories, if row $r$ has $K_r$ history-equivalence classes,
  let $D_r\in\{0,1\}^{|\Pi|\times K_r}$ have $(D_r)_{\pi k}=1$ exactly when
  policy $\pi$ belongs to history class $k$ on row $r$. Then
  \begin{equation}
  s_Z(q)=\frac{\sum_r w_r\lVert D_r^\top q\rVert_1}{\lVert q\rVert_1}.
  \label{eq:main-resolution}
  \end{equation}
  Writing $Y$ for the scored outcome and $P_{R,Z,Y}^{\pi}$ for the joint law
  induced by policy $\pi$, the
  declared history is \emph{score-sufficient for $q$} if there is one Markov
  kernel $L(\mathrm dy\mid r,z)$ such that
  $P_{R,Z,Y}^{\pi}=P_{R,Z}^{\pi}L$ for every
  $\pi\in\operatorname{supp}(q)$.
  \end{definition}

For a fixed set of fully specified deterministic update types, counts of each
type determine
the composite update map---and hence the final state from any initial
state---for every finite ordering exactly when the corresponding update maps
commute pairwise (Appendix Lemma~\ref{lem:count-sufficiency}). This result
concerns the final state: an intermediate answer can still depend on a prefix,
and a scalar total count is coarser than a count for each update type.

  \begin{theorem}[History-resolution certificate]
  \label{thm:trace-resolution-certificate}
  Assume a common initial state and coupled exogenous randomness (shared random
  inputs), the
  policy-independent row law in Definition~\ref{def:trace-resolution}, and that
  the declared history is score-sufficient for $q$ in the sense of that
  definition. Let $P_Y^{q^\pm}=P_{R,Z}^{q^\pm}L$ be the induced outcome laws and
  define
  $\Delta_e^{\rm exec}(q)=\int y\,\mathrm dP_Y^{q^+}(y)
  -\int y\,\mathrm dP_Y^{q^-}(y)$.
  For a measurable summary $h$, let $h_\#P$ denote the distribution obtained by
  applying $h$ to a draw from $P$.
  If $Y_e\in[a,b]$ almost surely, with $a<b$, then
  \begin{equation}
  |\Delta_e^{\rm exec}(q)|\leq(b-a)s_Z(q),
  \qquad
  \TV(h_\#P_{R,Z}^{q^+},h_\#P_{R,Z}^{q^-})\leq s_Z(q)
  \label{eq:bounded-resolution-effect}
  \end{equation}
  for every measurable summary $h(R,Z)$.
  \end{theorem}
  \noindent\emph{Full proof.}
  See Appendix~\ref{app:proof-trace-resolution-certificate}.

  History resolution lies in $[0,1]$. Zero means that the two weighted policy
  groups produce identical recorded-history mixtures and, under the theorem's
  assumptions, must have the same average score. One means that their recorded
  histories do not overlap, in which case the bound spans the full score range
  and does not constrain the effect magnitude. A complete derivation of
  Equation~\ref{eq:main-resolution} appears in
  Appendix~\ref{app:proof-resolution-formula}. The probability facts are
  classical; the tracing procedure makes the bound available before outcomes
  are examined.

  The appendix derives the minimum history resolution needed to support a
  specified score difference and extends the bound to a known amount of
  mismatch in the scoring process. These results require a defensible mismatch
  bound; they do not verify that the tracing rule captures every relevant path
  (Appendices~\ref{app:proof-resolution-floor}
  and~\ref{app:proof-kernel-mismatch}).

\subsection{Control Map records}

\begin{definition}[Control Map]
\label{def:control-map}
For each comparison, the Control Map keeps four kinds of evidence separate:
the recorded histories and how well they distinguish the update plans, the
estimated behavioral change and its uncertainty, acquisition, retention, and
locality, and measured cost. Missing or invalid evidence is reported as
\textsc{undefined}. To evaluate a conclusion in a new setting, we
apply the same record there under an explicit mapping of the tested update
plans. Appendix~\ref{sec:typed-evidence-record} gives the complete record
definitions.
\end{definition}
Distinguishable histories rule out complete aliasing, but they
neither determine effect direction nor replace the design assumptions needed
for effect estimation.

%% file: sections/03_update_reachability.tex
\section{Which Updates Reach Each Answer?}
\label{sec:execution}

Nominally different schedules are not always distinguishable at an answer:
short inputs permit fewer updates and therefore collapse more schedules to the
same pre-answer history. Longer inputs expose progressively more of the
schedule grid.

\subsection{Formulation}

For the Memory-as-Context (MAC) schedule experiment and an input containing
$L\geq1$ tokens, the number of complete writes available before scoring is
\begin{equation}
H(L)=\left\lfloor\frac{L+256\lfloor(L-1)/512\rfloor}{16}\right\rfloor,
\quad
a_p(H)=(\ind\{t\bmod p=0\})_{t=1}^{H}.
\label{eq:mac-horizon-v94}
\end{equation}
For $\Pi=\operatorname{Perm}(1,2,4,8)$ and
  $s=(p_1,\ldots,p_4)\in\Pi$, let
  $z_H(s)=(a_{p_1}(H),\ldots,a_{p_4}(H))$. Exact enumeration yields
  \begin{equation}
  |\{z_H(s):s\in\Pi\}|=
  \begin{cases}1,&H=0,\\4,&H=1,\\12,&H\in\{2,3\},\\24,&H\geq4.
  \end{cases}
  \label{eq:mac-staircase-v94}
  \end{equation}
  For a pair $q=\mathbf e_s-\mathbf e_{s'}$, where $\mathbf e_s$ is the
  policy-coordinate basis vector for schedule $s$,
  $s_Z(q)=1-\sum_rw_r\ind\{z_{r,s,e}=z_{r,s',e}\}$.
  The class count and pairwise-resolution identity are proved in
  Appendix~\ref{app:proof-mac-staircase}.

Appendix~\ref{app:execution-details} reports which layer--period comparisons
remain distinguishable when schedules collapse to the same history, together
with the resulting rank and conditioning diagnostics. The corresponding
estimability statement and proof appear in
Appendix~\ref{app:proof-working-basis-estimability}.

\subsection{Experimental protocol and results}

The history audit uses a fixed 24-layer, approximately 760M-parameter
  Memory-as-Context evaluator (MAC-760M) and enumerates all 24 assignments of
  periods $\{1,2,4,8\}$ to four six-layer depth bands on 27,072 rows from eight
  evaluation sets chosen before outcomes were examined. Policies share a nominal
  update count but differ in
  pre-answer opportunity. Appendix Figure~\ref{fig:appendix-resolution-profile}
  reports both the exact class staircase and the benchmark-specific row
  partition.

Of the 5,153 LAMBADA~\citep{paperno2016lambada} rows, 5,141 distinguish
  all schedules. For a row sampled uniformly from all 27,072 benchmark rows and
  a pair sampled uniformly from the 276 schedule pairs, the probability of
  identical histories is $0.1285$. A retrospective
  SocialIQA~\citep{sap2019socialiqa} check makes
  Theorem~\ref{thm:trace-resolution-certificate} concrete: each period-4 versus
  period-8 swap changes the declared pre-answer history on only one of 1,954
  rows. Thus $s_Z=1/1954$, any binary-accuracy contrast is bounded by $0.0512$
  percentage points, and all six observed contrasts are zero.

\paragraph{Live neural validation.}
  On a randomly initialized two-layer Llama model updated with
  LoRA~\citep{hu2022lora}, the tracing procedure recovers the value $s_Z=1/2$
  specified in advance from 16 rows and rejects all 16 injected trace
  corruptions; validation passes again after the original traces are restored.
  Each row performs updates A then B, but B precedes scoring in 6/8 rows for one
  policy and 2/8 for the other. This controlled test was fixed locally before
  model construction. It validates conformance of the recorded nonzero-gradient
  paths in this setup. It does not establish complete coverage of published
  editor internals or automatic discovery of every relevant path.

Finally, an apparent dose--locality curve, where dose is the number of updates
completed before the focal query, disappears after matching model, writer,
strength, initialization and order replicate, family, the focal item's position,
and complete event order: all 1,600 pairs have identical terminal model states
and locality scores.
After matching, data no longer support interpreting dose as the cause
of the terminal-locality pattern.

%% file: sections/04_behavioral_effects.tex
\section{Behavioral Effects of Control Choices}
\label{sec:effects}

For each policy comparison with sufficient support from the recorded-history
design, we report
the mean behavioral difference, its uncertainty, the inference unit, and when
the analysis was specified. Unless stated otherwise, 95\% ranges come from
paired family- or block-level reweighting of one fixed evaluation set; they
describe sensitivity to those units. In the editing studies, a family groups related rows for one target
fact. Appendix~\ref{app:effect-record} gives the formal record.

\subsection{Experimental protocol and results}
\begin{figure}[t]
    \centering
    \includegraphics[width=0.98\linewidth]
      {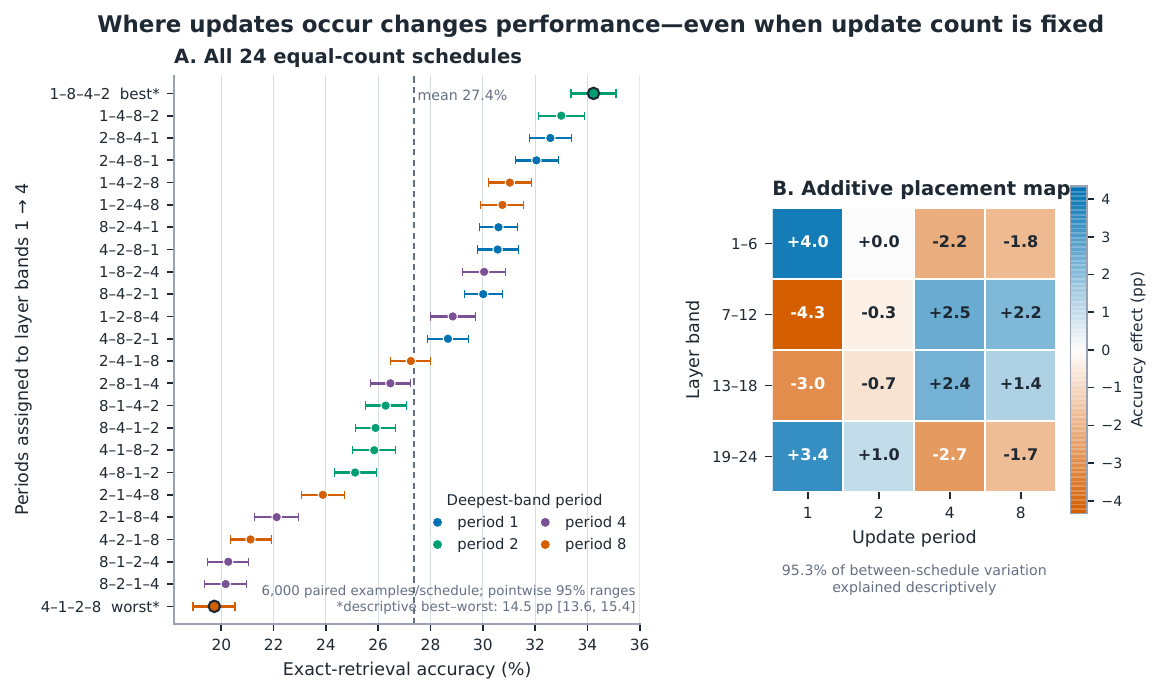}
    \caption{\textbf{Where updates occur changes performance even when update
    count is fixed.} All 24 assignments of periods $\{1,2,4,8\}$ to four ordered
    layer bands are evaluated on the same 6,000-example single-needle exact
    retrieval (Single-NIAH) evaluation set.
    Schedule codes list periods from the earliest to the latest layer band;
    whiskers are pointwise paired 95\% bootstrap ranges over the fixed
    evaluation set. The
    additive assignment map explains 95.3\% of between-schedule variation
    descriptively (leave-one-schedule-out $R^2=0.863$, root-mean-square error
    $=1.59$ percentage points). Best--worst labels and the decomposition are retrospective, so this
    is a local response surface rather than a universal schedule ranking.}
    \label{fig:placement-landscape}
  \end{figure}

Where and when updates occur can matter even when their total number is fixed,
but the resulting pattern is specific to the tested setting.
Figure~\ref{fig:placement-landscape} shows the complete equal-count placement
surface and its additive decomposition.

At this checkpoint, the fitted additive effects of frequent updates are
positive in the earliest and latest layer groups and negative in the middle
groups. We treat this as a local
placement--frequency interaction rather than a universal layer rule.
Table~\ref{tab:effect-slices} summarizes this placement result together with
the timing and dose comparisons.

\begin{table}[!htbp]
\centering
\footnotesize
\setlength{\tabcolsep}{1.5pt}
\caption{\textbf{Controlled comparisons of placement, timing, and dose.}
Each row changes one choice while preserving the listed controls; conclusions
remain specific to the tested setting.}
\label{tab:effect-slices}
\begin{tabularx}{\linewidth}{@{}>{\hsize=.55\hsize}Y
    >{\hsize=1.25\hsize}Y>{\hsize=1.45\hsize}Y
    >{\hsize=.75\hsize}Y@{}}
\toprule
Choice & Comparison and controls & Observed result & Supported takeaway \\
\midrule
Placement & 24 band--period assignments; one MAC-760M checkpoint; 6,000 paired Single-NIAH rows; equal prompt-only writes & Accuracy $19.733\%$--$34.233\%$; the band--period model descriptively explains $95.322\%$ of between-schedule variation & Placement matters in this task and checkpoint; \\
Timing & Same selected events and terminal actions; writes before versus after
  the focal query & Immediate-minus-deferred acquisition:
  $+0.1196,+0.0598,+0.4320,$ and $+0.1688$ nats across four GPT-2
  sizes~\citep{radford2019language}; Pythia-410M~\citep{biderman2023pythia}:
  $+0.1592$ nats, range $[-0.4484,+0.7667]$ & The estimated effect is positive
  in all four GPT-2 assays; the Pythia range crosses zero \\
Dose & Eight objectives, eight actions, 32 stochastic-gradient-descent steps, and terminal adapter--optimizer construction fixed & Across 18 deliberately varied LoRA settings, the best tested doses $1/2/4/8$ occur in $9/7/2/0$ settings; the best-dose prediction specified before outcomes is confirmed in 7/18 settings & The best tested dose varies across settings \\
\bottomrule
\end{tabularx}
\end{table}

%% file: sections/05_memory_feasibility.tex
\section{Feasibility of Acquired Memory}
\label{sec:feasibility}

\subsection{Decision rules}

We evaluate acquisition, retention, and locality separately using thresholds
specified for each study. We orient each decision metric so that larger values
are better; for locality, this means using the negative of measured damage. A
valid uncertainty range lies above its threshold
for \textsc{pass}, below it for \textsc{fail}, and across it for
\textsc{inconclusive}; an absent or invalid measurement is
\textsc{undefined}. The retention check also requires that at least 20\% of
edit pairs meet the immediate-learning condition (installation) and that at
least half of those installed edits are retained. The overall memory-quality
decision passes only when every required check passes, so strong acquisition
cannot compensate for failed retention or locality. Cost is reported
separately and becomes a requirement only when a budget was specified. The
complete decision rules appear in
Appendix~\ref{app:feasibility-gate-definitions}.

\subsection{Experimental protocol and results}

Immediate learning can overstate memory quality: a method may acquire the
requested information yet fail to retain it or preserve unrelated behavior.
The LoRA study measures query-time acquisition, terminal retention, and
terminal locality separately. Across six selected CounterFact~\citep{meng2022rome} settings,
acquisition passes 6/6 while retention and locality pass 0/6; the prespecified
four-setting preliminary screen passes in 0/4 and triggers its stopping rule
(Appendix~\ref{app:stage0-factual}). A
study of published editors exposes the same logical gap: ROME passes acquisition
and the specified retention checks at all 13 operating points, but none passes
the locality criterion or the full memory-quality requirement. A separate
shared-cohort comparison of dose-one writer profiles appears in
Appendix Figure~\ref{fig:writer-quality-radar}; acquisition--locality tradeoffs
and dose profiles appear in Appendix
Figure~\ref{fig:appendix-quality-dose}.

A separate 160-family positive-control evaluation set shows that the criteria are attainable. MEMIT-high has acquisition $8.30$ nats and
locality damage $2.08$ percentage points and passes the full CounterFact
decision rule. ROME-low has acquisition $7.52$ nats and damage $4.64$
percentage points, but its damage range $[3.08,6.42]$ crosses the threshold.
The MEMIT-minus-ROME damage contrast is $-2.56$ percentage points
$[-4.27,-1.06]$. Because the dose-four query is terminal, continuous retention
duplicates acquisition. Installation and retention among installed edits remain
separate checks; the latter can include subsequent edits, depending on the
focal edit's position.
In a separate shared-cohort eight-edit sequence, AlphaEdit meets all declared
memory-quality criteria at each of the four tested doses for these
setting-specific operating points.

The five-percentage-point locality boundary is a tolerance specified before
outcomes. Exact threshold analysis
(Appendix~\ref{app:v95-robustness}) places the CounterFact pass breakpoints at
$3.20$ percentage points for MEMIT-high and $6.42$ for ROME-low, versus
$52.73$ and $50.78$ on WikiRecent.
Relaxing locality cannot rescue LoRA because retention passes in 0/6 settings.
New deployments should specify setting-specific thresholds before examining
outcomes, using an explicit tolerated loss; every required component must pass.

%% file: sections/06_cross_setting_generalization.tex
\section{Generalization Across Models and Benchmarks}
\label{sec:transport}

\subsection{Comparing settings}

A \emph{bundle} is a fixed model and editor configuration. To ask whether a
result carries to another bundle or benchmark, we specify in advance how the
tested update plans correspond across the two settings and preserve which plan
is the positive side of the comparison. Within each setting, a valid
uncertainty range entirely above zero gives a positive sign, one entirely
below zero a negative sign, and one crossing zero an unresolved sign. These
endpoint-specific measurements are reported separately rather than combined
into a single utility. Data from one setting do not determine an unevaluated
setting without an additional invariance assumption, so our strongest claims
use direct evaluation in the new setting. Appendix~\ref{app:transport-formulation}
gives the formal definitions and the off-support construction.

\subsection{Results across two model bundles}

The relative performance of the tested editors varies across the evaluated
model--benchmark--question bundles; a source-setting ranking is therefore not
something that should be assumed to carry over.

\paragraph{GPT-2XL source.}
  With checkpoint, editor strengths, execution, dose, and estimator fixed, the
  MEMIT-minus-ROME acquisition contrast is positive on all three endpoints
  ($+0.7850,+0.5772,+1.2930$ nats). Its locality-success contrast changes from
  $+2.5625$ percentage points on CounterFact to $-7.5000$ on
  zsRE~\citep{levy2017zeroshot,zhang2024knowedit} and $-15.9180$ on
  WikiData-CounterFact~\citep{zhang2024knowedit}; the evaluation sets and
  questions differ.

Appendix Figure~\ref{fig:appendix-endpoint-trajectories} displays the complete
  endpoint trajectories with separate pointwise ranges for each fixed
  evaluation set.

\paragraph{Llama-3.2-3B follow-up.}
  Using Llama-3.2-3B~\citep{meta2024llama323b}, the analysis was fixed before
  outcomes. A later reconstruction of the recorded editor-call sequences gives
  terminal $s_Z=1$, although it neither inspects editor internals nor verifies
  the cross-model policy mapping. Ordered as CounterFact, zsRE, and
  WikiData-CounterFact, the predicted $(+,-,-)$ locality pattern does not recur:
  the result is $(+,+,?)$. Acquisition favors ROME on all three endpoints
  $(-0.3017,-0.2116,-0.2247)$ nats; both editors fail every locality requirement.

 \paragraph{Prospective WikiRecent.}
  The only cross-setting workflow for which the policy mapping, tracing
  procedure, outcome measures, and decision rules were all fixed before setup
  validation or outcomes were inspected uses 64
  WikiRecent~\citep{zhang2024knowedit} families, four cyclic orders, and two
  GPT-2XL policies. All 512 editor-call sequences match their specifications.
  MEMIT's gain after four edits relative to the reset baseline exceeds ROME's by
  $+0.2142$ nats per target token on canonical prompts and $+0.7178$ on a
  rephrase, satisfying the prespecified requirement that MEMIT's mean gain
  exceed ROME's. Yet both editors fail the two-probe locality requirement
  ($-43.36$ and $-41.41$ percentage points), so the overall evidence requirement
  fails.

\paragraph{Alternative outcome measures.}
With rows, model states, probes, and bootstrap draws fixed, binary success,
full-vocabulary Kullback--Leibler divergence, and greedy-token change all favor
MEMIT at the source setting when each measure is oriented so that positive
means better performance. Complete results and scope appear in
Appendix~\ref{app:v95-robustness}.

\subsection{What the combined evidence supports}
\label{sec:discussion}

The observed rank changes and locality failures show why relative performance
and absolute memory quality must be reported separately. Update histories
establish what differed before scoring; evaluation in a new setting establishes
whether a result recurs there. Appendix Table~\ref{tab:decision-replacements}
summarizes four common shortcuts and the corresponding evaluation alternatives.

\paragraph{Limitations.}
The guarantees depend on the stated tracing rule. Our checks verify that the
recorded event sequences follow their specifications. If an omitted path
invalidates score sufficiency, Theorem~\ref{thm:trace-resolution-certificate}
does not apply. The cross-setting grid also changes the model, benchmark, and
test questions together, so it
demonstrates heterogeneity without identifying its cause. Detailed execution
checks, robustness bounds, measurement times, and reproducibility information
appear in
Appendices~\ref{app:v95-robustness} and~\ref{app:artifact}.

%% file: sections/07_conclusion.tex
\section{Conclusion}
\label{sec:conclusion}

Evaluations of test-time adaptation should report the ordered updates completed
before each scored answer, the update mechanism and configuration, and the
measurement times. Acquisition, retention after later updates, and locality
should have separate results and uncertainty estimates; costs should be reported
alongside them. Comparisons in a new
model or benchmark should state how update plans correspond and measure outcomes
there directly. This reporting practice makes visible the scheduling differences
and memory-quality failures that a total update count or an immediate score can
hide.

%% file: sections/backmatter/statements.tex
\clearpage

\subsection*{AI use statement}

Generative AI tools assisted with literature discovery; manuscript drafting and
editing; research code development, debugging, and auditing; data cleaning and
formatting; construction of synthetic evaluation panels; preparation of
analyses, tables, and figures; and development and checking of mathematical
arguments, statistical claims, and textual claims. The authors determined the
research questions, experimental designs, estimands, decision rules, and final
interpretations. All AI-assisted outputs were reviewed. Claim-bearing
computations were tested or independently reproduced during the research.
The public bundle supports the subset of recomputations described below.
AI-generated suggestions were not treated as evidence. The authors take full
responsibility for the paper.

\subsection*{Ethics statement}

This work evaluates failure modes in model memory and editing; it involves no
human participants and no deployment. Such mechanisms can store false or
sensitive associations and damage unrelated behavior, so we report locality
and retention failures rather than treating target recall as sufficient. We
restrict claims to tested systems. Artifact release is subject to privacy,
redistribution rights, and third-party license checks.

\subsection*{Reproducibility statement}

The paper specifies estimands, units, schedules, decision rules, tracing rules,
and the scope of its claims. The reproducibility bundle includes code for
finite deterministic histories, figures and tables generated from recorded
aggregates, sanitized WikiRecent sufficient statistics, and validation scripts.
Offline checks recompute the structural outputs and complete WikiRecent record.
For one CPU study, the source and protocol were recorded before execution;
a separate implementation checks its history mixtures and corruption controls.
A second CPU check recomputes the finite omitted-path example,
decision-threshold breakpoints, and structural LoRA history tests.
The bundle omits legacy family-level measurements, so readers cannot
independently recompute their reported uncertainty intervals from it. Raw data,
model checkpoints, and third-party editor internals are also excluded.
We plan to release redistributable author-created code, protocols, aggregates,
and validators. Third-party data and checkpoints retain their original licenses.

%% file: sections/appendices/audit_and_empirical_details.tex
\setlength{\intextsep}{8pt plus 2pt minus 2pt}

\section{Evidence records and mathematical scope}
\label{app:audit-contract}

Section~\ref{app:audit-contract} defines the evidence records and presents the
resolution results, adapter checks, and gate sensitivity analyses.
Section~\ref{app:execution-details} reports execution and effect details;
Section~\ref{app:feasibility-details} reports absolute feasibility;
Section~\ref{app:transport-details} reports comparisons across endpoints;
Section~\ref{app:artifact-boundary-v93} describes the reproducibility boundary;
and Section~\ref{app:formal-proofs} contains the proofs.

The operational schematic appears as Figure~\ref{fig:controller-overview} in
the main paper. Here LoRA denotes low-rank adaptation and SGD denotes
stochastic gradient descent.

\begin{table}[H]
\centering
\caption{\textbf{Studies and supporting evidence.} Each row identifies the
study, analysis timing, inference or enumeration unit, and supporting section.
Assays answer separate questions and are reported separately. ``Frozen'' means
fixed and recorded before the named outcome was accessed; these protocols were
not publicly preregistered.}
\label{tab:study-design}
\scriptsize
\begingroup
\setlength{\tabcolsep}{3pt}
\renewcommand{\arraystretch}{0.96}
\begin{tabularx}{\textwidth}{@{}>{\raggedright\arraybackslash}p{0.19\textwidth}>{\raggedright\arraybackslash}p{0.20\textwidth}>{\raggedright\arraybackslash}p{0.18\textwidth}>{\raggedright\arraybackslash}p{0.20\textwidth}Y@{}}
\toprule
Claim / audit question & Study & Timing & Unit and scale & Location \\
\midrule
Equal counts need not identify equal executed treatments & MAC-760M trace atlas; 24 depth--cadence schedules on eight panels & Outcome-blind exact enumeration & 27,072 benchmark rows; 24 policies per row & Section~\ref{app:execution-details}; Table~\ref{tab:atlas-v93} \\
Layer placement can change accuracy at fixed update count & MAC-760M placement surface on Single-NIAH & One fixed checkpoint; additive summary descriptive & 6,000 paired rows; all 24 schedules & Figure~\ref{fig:placement-landscape}; Section~\ref{app:placement-response} \\
Order can masquerade as dose--locality structure & GPT-2XL LoRA complete-order correction & Post-hoc deterministic correction & 1,600 order-matched pairs & Section~\ref{app:execution-details} \\
Query-relative phase can change acquisition & GPT-2 family and Pythia-410M phase studies & Per-model frozen gates; cross-model synthesis descriptive & Four LoRA initializations per model & Section~\ref{app:execution-details} \\
The best tested dose varies across settings & LoRA/SGD four-dose grids & Purposive, partly adaptive setting collection; within-setting calls frozen & 18 settings; four adapter initializations per setting & Section~\ref{app:dose-summary} \\
Acquisition need not imply feasible memory & LoRA and ROME grids; fixed MEMIT, ROME, and AlphaEdit controls & Cohort-specific frozen or declared timing & Family/block resampling within each assay & Table~\ref{tab:feasibility}; Sections~\ref{app:stage0-factual}--\ref{app:additional-editor-breadth} \\
Observed editor ordering varies across evaluated bundles & GPT-2XL and Llama-3.2-3B across CounterFact, zsRE, and WikiData-CF & Endpoints separately frozen; Llama follow-up fixed before Llama outcomes & 160/160/128 families; four orders per bundle & Sections~\ref{app:endpoint-transport} and~\ref{app:llama-transport} \\
Strict prospective end-to-end verdict & GPT-2XL MEMIT-high/ROME-low on WikiRecent & Full workflow frozen before preflight or outcomes & 64 families; four Latin orders; 512 policy rows & Section~\ref{app:wikirecent-prospective} \\
Whether the tracing procedure detects changes to live neural execution & Random two-layer Llama with PEFT-LoRA and two SGD writes & Source and protocol recorded locally before model construction; no external preregistration & 16 policy rows; two history classes; 16 corruptions plus restoration & Section~\ref{app:v95-robustness} \\
How conclusions vary with hidden paths, adapter scope, and gate tolerance & Finite hidden-cache witness; adapter-conformance audits; exact breakpoint analysis & Deterministic post-hoc robustness audits & Exact finite witness, recorded traces, and frozen gate records & Section~\ref{app:v95-robustness} \\
\bottomrule
\end{tabularx}
\endgroup
\end{table}

\subsection{Secondary resolution guarantees}

\begin{lemma}[When counts are sufficient]
\label{lem:count-sufficiency}
Fix a finite atom alphabet $\mathcal A$ and a state space $\mathcal X$. For
each $a\in\mathcal A$, let $T_a:\mathcal X\to\mathcal X$ be a deterministic
self-map. For a word $\Gamma=a_1\cdots a_T$, set
$F(\Gamma)=T_{a_T}\circ\cdots\circ T_{a_1}$ and let $\nu(\Gamma)$ be its
atom-count vector, with $F(\varnothing)=\mathrm{id}_{\mathcal X}$. Over all
finite words, there exists
$Q:\mathbb N_0^{|\mathcal A|}\to
\{f\mid f:\mathcal X\to\mathcal X\}$ such that
\begin{equation}
F(\Gamma)=Q(\nu(\Gamma))\quad\text{for every }\Gamma
\label{eq:count-factorization-v94}
\end{equation}
if and only if the maps $T_a$ commute pairwise.
\end{lemma}
The proof appears in Appendix~\ref{app:proof-count-sufficiency}.

\begin{remark}[Final-state scope]
\label{rem:final-state-scope}
An intermediate query can depend on a prefix even when final maps commute, and
scalar total count is coarser than $\nu$. On a restricted word catalogue,
counts may also suffice accidentally without global commutativity.
\end{remark}

\begin{corollary}[Resolution floor for a target effect]
\label{cor:resolution-floor}
Under Theorem~\ref{thm:trace-resolution-certificate} with $a<b$, any claim that
the realized assignment effect satisfies
$|\Delta_e^{\rm exec}(q)|\geq\delta>0$ requires
$s_Z(q)\geq\delta/(b-a)$. Thus $s_Z(q)<\delta/(b-a)$ establishes before
outcome access that the target effect is unattainable under the theorem's
assumptions.
\end{corollary}
The proof appears in Appendix~\ref{app:proof-resolution-floor}.

\begin{proposition}[Robustness to conditional-kernel mismatch]
\label{prop:kernel-mismatch}
Let $X=(R,Z)$, let $K_+(\mathrm dy\mid x)$ and $K_-(\mathrm dy\mid x)$ be
the actual outcome kernels in the two contrast arms, and let
$L(\mathrm dy\mid x)$ be any common reference kernel. Define
$P_Y^{q^\pm}=P_X^{q^\pm}K_\pm$ and
$\epsilon_\pm=\int\TV(K_\pm(\cdot\mid x),L(\cdot\mid x))
\,\mathrm dP_X^{q^\pm}(x)$. Let $\Delta_e^{\rm exec}(q)$ be the difference
between the means under these two output laws. If $Y_e\in[a,b]$ almost surely
under both laws, with $a<b$, and the two pointwise total-variation functions
in the definition of $\epsilon_\pm$ are measurable, then
\begin{equation}
\TV(P_Y^{q^+},P_Y^{q^-})
\leq \min\{1,s_Z(q)+\epsilon_++\epsilon_-\},\qquad
|\Delta_e^{\rm exec}(q)|
\leq (b-a)\min\{1,s_Z(q)+\epsilon_++\epsilon_-\}.
\label{eq:approximate-trace-sufficiency-v95}
\end{equation}
\end{proposition}
The proof appears in Appendix~\ref{app:proof-kernel-mismatch}. This extension
is informative only when an external bound on $\epsilon_+$ and $\epsilon_-$ is
defensible; it does not verify that the tracing rule is complete. The finite
hidden-cache experiment in Appendix~\ref{app:v95-robustness} illustrates the
boundary exactly.

\input{sections/appendices/evidence_contract.tex}
\input{sections/appendices/trace_resolution_theory.tex}
\input{sections/appendices/adapter_robustness_and_gate_sensitivity.tex}

\section{Execution and effect details}
\label{app:execution-details}

\begin{proposition}[Working-basis estimability]
\label{prop:working-basis-estimability}
Let $P_r=D_r(D_r^\top D_r)^{\dagger}D_r^\top$ be the orthogonal projector
that averages policy coordinates within row $r$'s executed-history classes.
Let $U\in\R^{|\Pi|\times d}$ be a centered design basis,
$\mathbf1^\top U=0$, and let
$\mu=\alpha\mathbf1+U\beta$ be the working vector of policy means. Define
$I_U=U^\top(\sum_rw_rP_r)U$. A functional $a^\top\beta$ is structurally
estimable when it has the same value for every two parameter pairs producing
the same projected means $P_r\mu$ on all rows with $w_r>0$. Then
$a^\top\beta$ is structurally estimable if and only if
$a\in\operatorname{range}(I_U)$.
\end{proposition}
The proof appears in Appendix~\ref{app:proof-working-basis-estimability}.

\subsection{Outcome-blind trace atlas}

The atlas uses the fixed MAC-760M evaluator, 24 permutations of periods
$\{1,2,4,8\}$ over four ordered six-layer bands, and eight complete benchmark
panels. If $L$ is the tokenizer length of the write-enabled common prompt,
16-token native chunks, 512-token data segments, and 256 inserted memory
tokens before each segment after the first give
\begin{equation}
 H(L)=\left\lfloor
 \frac{L+256\left\lfloor(L-1)/512\right\rfloor}{16}
 \right\rfloor .
 \label{eq:closed-form-transducer-v93}
\end{equation}
For period $p$, the phase-zero trace is
$a_p(H)=(\mathbf 1\{t\bmod p=0\})_{t=1}^{H}$. Direct enumeration gives
1, 4, 12, and 24 effective schedule classes for $H=0$, $H=1$,
$H\in\{2,3\}$, and $H\geq4$. Candidate and target tokens do not enter $H$
because writes are disabled before they are scored.

\begin{figure}[H]
  \centering
  \includegraphics[width=0.98\textwidth]{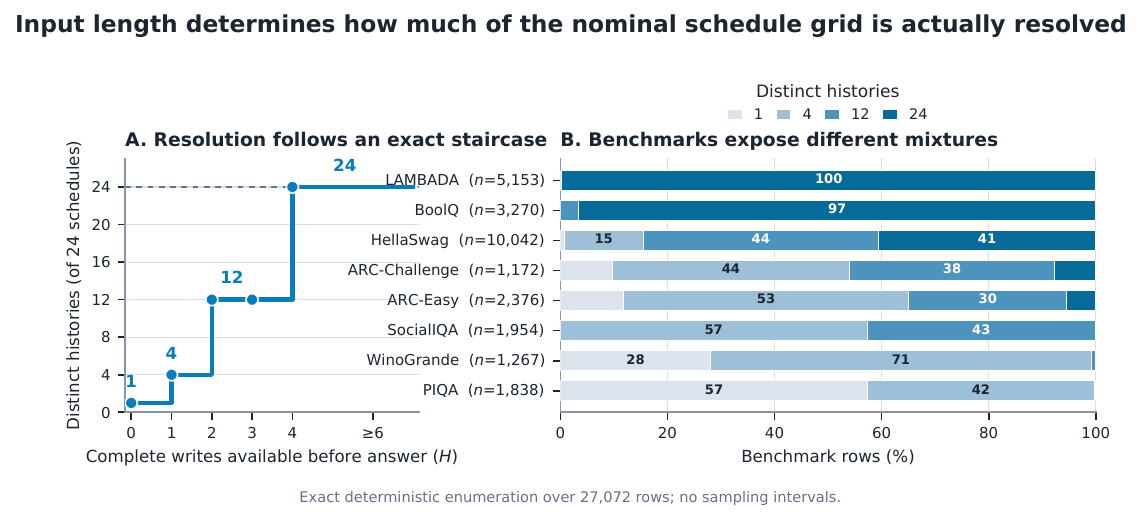}
  \caption{\textbf{Resolution depends on the pre-answer write horizon.}
  Exact enumeration yields 1, 4, 12, or 24 distinct histories as the number of
  complete writes available before an answer increases. The right panel shows
  the resulting mixture for each benchmark, with each bar normalized by that
  benchmark's row count. The calculation covers all 27,072 rows and is
  deterministic, so no sampling intervals apply.}
  \label{fig:appendix-resolution-profile}
\end{figure}

The eight evaluation panels are
  SocialIQA~\citep{sap2019socialiqa},
  BoolQ~\citep{clark2019boolq},
  LAMBADA~\citep{paperno2016lambada},
  PIQA~\citep{bisk2020piqa},
  WinoGrande~\citep{sakaguchi2020winogrande},
  HellaSwag~\citep{zellers2019hellaswag}, and the ARC Easy and Challenge
  sets~\citep{clark2018arc}.

\begin{table}[H]
\centering
\caption{Complete execution atlas. ``Pair alias'' is the mean fraction of 276
nominal schedule pairs with equal rowwise traces. ``Full rows'' is the
percentage of rows that distinguish all 24 policies. Rank and condition number
$\kappa$ refer to the centered nine-dimensional band--period working basis and
diagnose design support, not causal identification or sampling power.}
\label{tab:atlas-v93}
\scriptsize
\begin{tabular}{lrrrrr}
\toprule
Panel & Rows & Pair alias & Full rows & Rank/9 & $\kappa(I_U)$ \\
\midrule
SocialIQA & 1,954 & 14.31\% & 0.05\% (1 row) & 9 & 1954.000 \\
BoolQ & 3,270 & 0.15\% & 96.61\% & 9 & 1.035 \\
LAMBADA & 5,153 & 0.01\% & 99.77\% & 9 & 1.002 \\
PIQA & 1,838 & 66.60\% & 0.00\% & 6 & --- \\
WinoGrande & 1,267 & 43.58\% & 0.00\% & 6 & --- \\
HellaSwag & 10,042 & 5.84\% & 40.60\% & 9 & 2.445 \\
ARC-Easy & 2,376 & 24.65\% & 5.39\% & 9 & 16.375 \\
ARC-Challenge & 1,172 & 21.02\% & 7.68\% & 9 & 11.756 \\
\bottomrule
\end{tabular}
\end{table}

Across all 27,072 rows, the mean collision probability over a uniformly
chosen row and unordered policy pair is
$960{,}264/7{,}471{,}872=0.128517$. SocialIQA illustrates why global
injectivity is insufficient. Let $P_z$ project the 24-policy response vector
onto values constant within row $z$'s trace classes, let
$R=1954^{-1}\sum_zP_z$, and let $U$ be the orthonormal centered
nine-dimensional band--period contrast basis. The smallest eigenvalue of
$U^\top R U$ is only $1/1954$. The atlas calculation passes
184 deterministic checks without reading model outcomes.
Condition numbers are reported only for full-rank panels; a dash denotes rank
deficiency. They describe support geometry in the declared working basis. The
27,072 rows are benchmark rows, not independent training trajectories or an
experimental sample-size claim.

In a retrospective SocialIQA outcome check, six period-4/period-8 schedule
swaps differ in declared pre-answer history on only the single fully resolving
row, giving $s_Z=1/1954$ for each pair. Their logits differ only on that row,
and their predictions and accuracies are identical. These six swaps therefore
do not support a broad schedule effect on SocialIQA; a separately named pair in
the same panel is fully resolved.

\subsection{Complete-order locality correction}

The four-event editor grids index dose through query placement while terminal
state follows the complete event sequence. Dose-indexed Latin-square cells can
therefore differ in focal rank and event order. The corrected comparison holds
editor, strength, replicate, family, and the complete ordered case
IDs fixed. It covers 160 MEMIT, 480 ROME, and 960 LoRA cross-dose pairs.
All 1,600 pairs reach the same terminal parameters and give identical
locality. The earlier terminal dose--locality curve therefore mixed event
orders. Acquisition is measured at query time, before the shared terminal state.

\subsection{Immediate-versus-deferred effect}

The phase comparison holds selected objectives, total actions, optimizer
steps, and terminal state fixed while moving the same writes across the focal
query. Immediate and deferred policies each execute 772 selected actions and
3,088 optimizer steps per endpoint. The held-out immediate-minus-deferred
means are $+0.119608$, $+0.059816$, $+0.432027$, and $+0.168832$ nats for
GPT-2 Small, Medium, Large, and XL; their separate Student-$t_3$ intervals are
$[+0.096519,+0.142698]$, $[+0.053095,+0.066537]$,
$[+0.294741,+0.569313]$, and $[+0.140598,+0.197066]$. Each model has four
LoRA-initialization units and 4/4 positive signs; under an independent and
identically distributed (i.i.d.) symmetric-sign null, the exact one-sided probability is
$1/16$, so each frozen gate relies on its stated
parametric interval.

The prospective Pythia-410M confirmation returns $+0.1592$ nats with
Student-$t_3$ interval $[-0.4484,+0.7667]$ and only 3/4 positive signs.
Execution checks pass, but the interval includes zero, so the study does not
confirm the effect. The five studies share one constructed 128-row association
panel; the four GPT-2 sizes are analyzed separately and do not become an
$n=16$ scale curve.

\subsection{Equal-count placement response}
\label{app:placement-response}

The placement study evaluates all 24 assignments of periods
$\{1,2,4,8\}$ to four ordered six-layer bands on the same 6,000 Single-NIAH
rows. Accuracy ranges from $19.733\%$ for schedule $(4,1,2,8)$ to $34.233\%$
for $(1,8,4,2)$. The additive band--period description accounts for
$95.322\%$ of between-schedule variation; leave-one-schedule-out prediction
has $R^2=0.863$ and root-mean-square error $1.592$ percentage points. These are
descriptive results for one checkpoint and evaluation set. The full surface is
shown in Figure~\ref{fig:placement-landscape}.

\subsection{Selected-setting dose summary}
\label{app:dose-summary}

Across 18 purposively and partly adaptively assembled LoRA/SGD settings, the
empirical peak at doses $1/2/4/8$ occurs in $9/7/2/0$ settings, respectively,
and the best-dose prediction specified before outcomes is confirmed in 7 of 18
settings. Each setting
uses the same four-dose grid and four adapter initializations, but the models,
checkpoints, writers, and timing classes differ. The counts are therefore a
descriptive summary of the selected grid, not an estimate of population
prevalence or a universal dose rule.

\subsection{Effect record and uncertainty}
\label{app:effect-record}

For setting $c$ and metric $m$, let $P_{c,m}$ be the evaluation distribution
over examples $r$ and exogenous randomness $\xi$, and let
$Y^m_{r,c}(\pi,\xi)$ be the resulting score. Define
$\mu_c^m(\pi)=\mathbb E_{(r,\xi)\sim P_{c,m}}
[Y^m_{r,c}(\pi,\xi)]$. For a policy comparison $q$ with sufficient support
from the recorded-history design, the effect record is
\begin{equation}
\Delta_c^m(q)=\frac{2}{\lVert q\rVert_1}
\sum_{\pi\in\Pi}q_\pi\mu_c^m(\pi),
\qquad
\mathsf{Eff}_c^m(q)=(\widehat\Delta_c^m,I_c^m,\mathcal U_c,\tau_c),
\label{eq:effect-record-v94}
\end{equation}
where $I_c^m$ is the uncertainty summary, $\mathcal U_c$ the inference unit,
and $\tau_c$ the analysis-timing label. The effect is evaluated only for a
comparison supported by the recorded histories and does not replace the
separate memory-quality checks. Unless stated otherwise, the reported 95\%
ranges apply the paired family- or block-level reweighting draws to one fixed
evaluation set; they describe sensitivity to those units rather than
population coverage.

\section{Absolute feasibility evidence}
\label{app:feasibility-details}

The following subsections report the gate values, inference units, and analysis
timing for the main-text feasibility results. Each cohort is reported separately.

\subsection{Feasibility decision rules}
\label{app:feasibility-gate-definitions}

Let $I_A=[\ell_A,u_A]$, $I_R=[\ell_R,u_R]$, and
$I_L=[\ell_L,u_L]$ be the finite-panel ranges for acquisition, retention, and
locality success, respectively. For a valid observed range requiring evidence
above threshold $\tau$, define
\begin{equation}
\mathsf S_\tau([\ell,u])=
\begin{cases}
\textsc{pass},&\ell>\tau,\\
\textsc{fail},&u<\tau,\\
\textsc{inconclusive},&\ell\leq\tau\leq u,\\
\textsc{undefined},&\text{the required record is absent or invalid}.
\end{cases}
\label{eq:feasibility-gates-v94}
\end{equation}
Let $\mathsf S_\tau^{\rm pt}(v)$ be \textsc{pass} if $v\geq\tau$ and
\textsc{fail} otherwise, with missing or invalid values \textsc{undefined}.
For a point--range record, let $\mathsf J_\tau(v,[\ell,u])$ be \textsc{pass}
if $v\geq\tau$ and $\ell>\tau$, \textsc{fail} if $v<\tau$ and $u<\tau$,
and \textsc{inconclusive} otherwise; it is \textsc{undefined} if either input
is absent or invalid. These definitions preserve each study's pass criterion.

Let $\widehat p_{\rm inst}$ be the fraction of edit pairs meeting the declared
immediate-learning condition, $\widehat p_{\rm keep}$ the fraction retained
among those installed edits, and $\widehat L$ the point estimate of locality
success. Using the status conjunction $\bigwedge_{\mathcal S}$ with precedence
\textsc{fail}$>$\textsc{undefined}$>$\textsc{inconclusive}$>$\textsc{pass},
and retaining each study's thresholds, the reporting records are
\begin{equation}
\begin{aligned}
S_A&=\mathsf S_0(I_A),\\
S_R&=\bigwedge_{\mathcal S}\!\left(\mathsf S_0(I_R),
 \mathsf S_{0.20}^{\rm pt}(\widehat p_{\rm inst}),
 \mathsf S_{0.50}^{\rm pt}(\widehat p_{\rm keep})\right),\\
S_L&=\mathsf J_{-5}(\widehat L,I_L),\qquad
S_{\rm full}=\bigwedge_{\mathcal S}(S_A,S_R,S_L).
\end{aligned}
\label{eq:feasibility-status-v94}
\end{equation}
The retention-among-installs field is \textsc{undefined} when no installation
occurs. Cost is reported separately and becomes a gate only under a declared
budget.

\begin{table}[H]
\centering
\caption{\textbf{Acquisition cannot stand in for feasibility.} Rows summarize
separate assays with their own counts, horizons, cohorts, and inference units.
A full pass requires every behavioral condition specified for that study;
cost is reported separately because no common budget was declared.}
\label{tab:feasibility}
\scriptsize
\begin{tabularx}{\textwidth}{p{0.29\textwidth}p{0.13\textwidth}p{0.13\textwidth}p{0.13\textwidth}Y}
\toprule
Assay & Acquisition & Retention & Locality & Full gate \\
\midrule
LoRA, six selected CounterFact settings & 6/6 \textsc{pass} & 0/6 \textsc{pass} & 0/6 \textsc{pass} & 0/6 \textsc{pass} \\
ROME, 13 CounterFact operating points & 13/13 \textsc{pass} & 13/13 \textsc{pass} & 0/13 \textsc{pass} & 0/13 \textsc{pass} \\
MEMIT-high, pilot-disjoint 160-family CounterFact point & \textsc{pass} & \textsc{pass} & \textsc{pass} & \textsc{pass} \\
ROME-low, same 160-family point & \textsc{pass} & \textsc{pass} & \textsc{inconclusive} & \textsc{inconclusive} \\
AlphaEdit, shared-eight-event CounterFact doses & 4/4 \textsc{pass} & 4/4 \textsc{pass} & 4/4 \textsc{pass} & 4/4 \textsc{pass} \\
\bottomrule
\end{tabularx}
\end{table}

\begin{figure}[!htbp]
  \centering
  \includegraphics[width=0.99\textwidth]
    {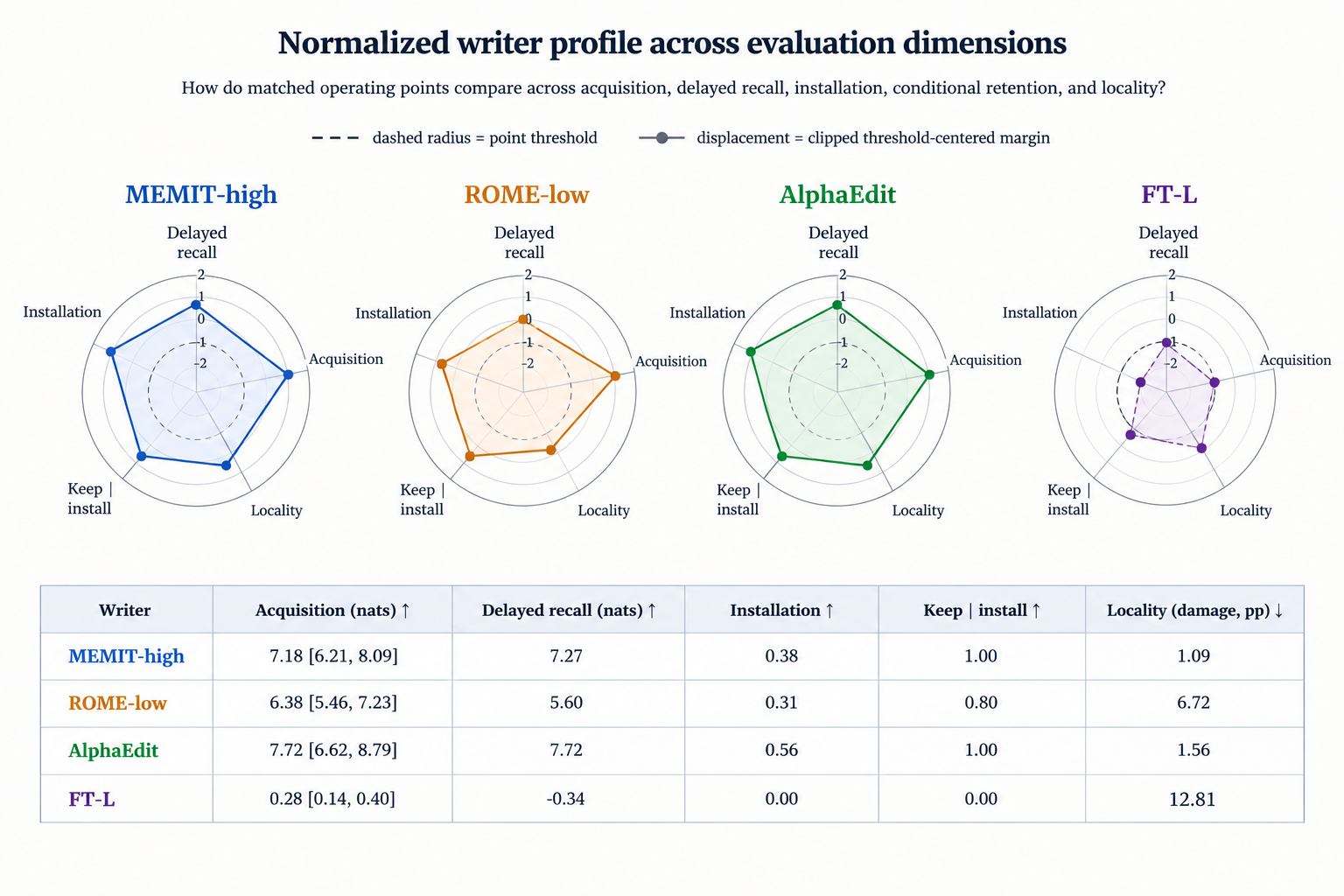}
  \caption{\textbf{Dose-one writer profiles differ across evaluation
  dimensions.} The threshold-centered display uses a shared eight-event
  CounterFact cohort with 16 disjoint family-pair blocks per writer
  (32 focal orientations and four event orders; 128 rows per writer). Outward is
  favorable, the dashed radius marks each axis's threshold, and the embedded
  table reports values on their native scales. Dose one indexes query-time
  acquisition; delayed recall and locality are measured after the shared
  eight-event sequence. Formal acquisition, delayed-recall, and locality
  decisions use recorded intervals; installation and keep-among-installs use
  point thresholds, and those intervals are not displayed on the spokes. The
  axes are separate requirements, so polygon area is not an overall utility or
  an editor-family ranking. FT-L denotes the constrained fine-tuning baseline.
  Because FT-L has no installations, its displayed 0.00 at
  Keep$\mid$install is a layout placeholder; the formal conditional-retention
  field is \textsc{undefined}.}
  \label{fig:writer-quality-radar}
\end{figure}

\begin{figure}[!htbp]
  \centering
  \includegraphics[width=0.98\textwidth]{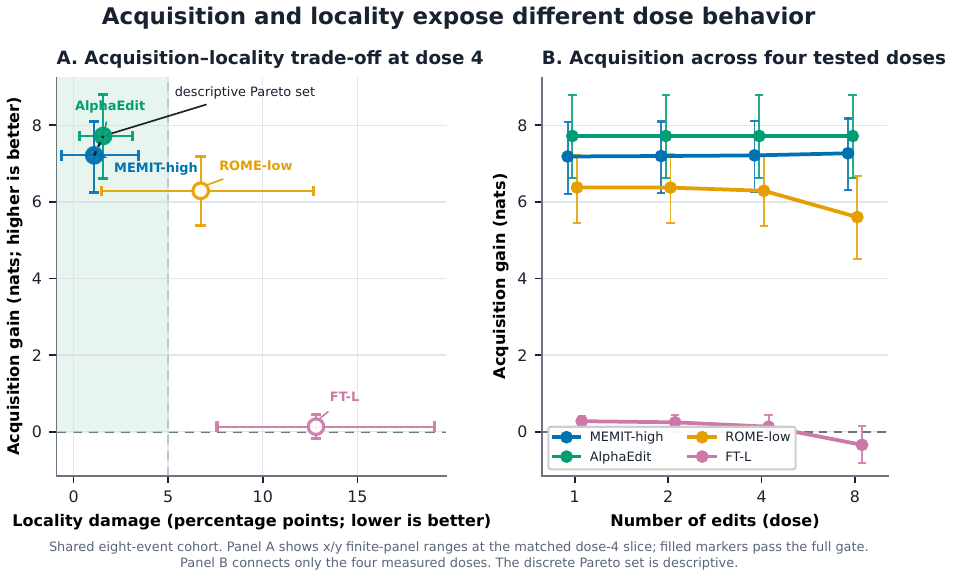}
  \caption{\textbf{Acquisition and locality expose different dose behavior on
  the shared eight-event cohort.} Panel A juxtaposes query-time acquisition
  with terminal locality at the matched dose-four slice and shows separate
  marginal finite-panel ranges; filled markers pass the full declared gate.
  Panel B shows query-time acquisition at the four measured doses
  $\{1,2,4,8\}$ only. The Pareto annotation and connecting lines are
  descriptive: they neither identify a causal dose trade-off, interpolate
  between doses, nor define an uncertainty-adjusted frontier.}
  \label{fig:appendix-quality-dose}
\end{figure}

\input{sections/appendices/matched_factual_recall_stage0.tex}
\input{sections/appendices/factual_recall_family_extension.tex}
\input{sections/appendices/editor_schedule_grids.tex}
\input{sections/appendices/memit_eight_event_study.tex}
\input{sections/appendices/fresh_editor_confirmation.tex}
\input{sections/appendices/frozen_160_family_extension.tex}
\input{sections/appendices/adaptive_rome_facet_sensitivity.tex}
\input{sections/appendices/additional_editor_dose_study.tex}

\section{Endpoint transport}
\label{app:transport-details}

The GPT-2XL and Llama comparisons report each endpoint's estimand, pointwise
reweighting range, absolute gates, tracing scope, and analysis timing.
In tables, ``WikiData-CF'' denotes WikiData-CounterFact.

\subsection{Formal comparison across settings}
\label{app:transport-formulation}

Fix a reference contrast $q_0$. For each endpoint--bundle cell
$j=(c_j,e_j)\in\mathcal J$, a prespecified correspondence
$\iota_j:\operatorname{supp}(q_0)\to\Pi_{c_j}$ induces the cell contrast
\begin{equation}
q_j(\pi)=\sum_{\pi_0:\iota_j(\pi_0)=\pi}q_0(\pi_0).
\label{eq:transport-pushforward-v94}
\end{equation}
The comparison is \textsc{undefined} unless each $q_j$ is nonzero and
preserves the declared positive-versus-negative orientation. For the
endpoint-specific metric $m_j$, let
$\widehat\Delta_j^{m_j}(q_j)=2\lVert q_j\rVert_1^{-1}
\sum_{\pi\in\Pi_{c_j}}q_{j,\pi}\widehat\mu_j^{m_j}(\pi)$ have valid range
$I_j=[\ell_j,u_j]$, and set
\begin{equation}
\rho_j=\begin{cases}+,&\ell_j>0,\\-,&u_j<0,\\?,&0\in I_j,
\end{cases}
\qquad
\boldsymbol\sigma_{\mathcal J}^{\mathbf m}
(q_0;\{\iota_j\}_{j\in\mathcal J})=(\rho_j)_{j\in\mathcal J}.
\label{eq:transport-signature-v94}
\end{equation}
A $?$ is unresolved, not a tie; endpoint-specific metrics are not pooled into
a common utility. A missing or invalid range makes $\rho_j$ undefined.

\begin{remark}[Off-support comparison]
\label{rem:off-support-transport}
Let $\mathcal H$ be the evaluated setting--policy--endpoint support and
$x^\dagger\notin\mathcal H$. Over unrestricted bounded response surfaces whose
outcome range contains at least two values, the record on $\mathcal H$ does not
identify the response at $x^\dagger$: two surfaces can agree everywhere on
$\mathcal H$ and differ at $x^\dagger$. A source ordering therefore needs a
stated invariance condition or new target evidence. Likewise, resolved signs
$\rho_j=+$ and $\rho_{j'}=-$ reject a constant-sign hypothesis for the mapped
contrast, while unresolved cells remain explicit. Because the measurements
are endpoint-specific, this is not a cross-instrument utility ranking.
\end{remark}
The construction is proved in Appendix~\ref{app:proof-off-support}.

\begin{figure}[H]
  \centering
  \includegraphics[width=0.98\textwidth]{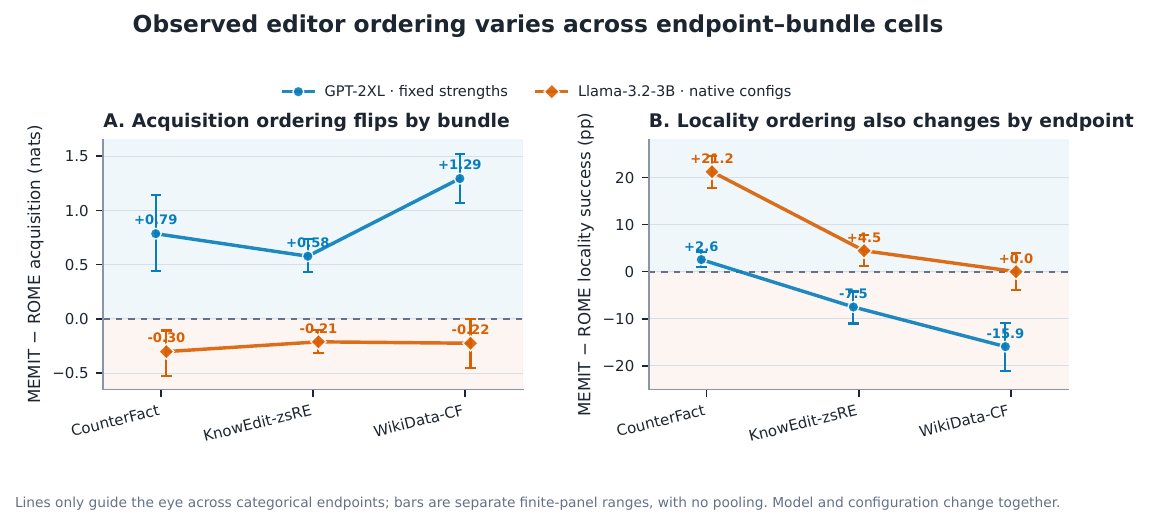}
  \caption{\textbf{Observed editor contrasts vary across evaluated
  endpoint--bundle cells.} Values are MEMIT-minus-ROME contrasts with separate
  pointwise paired-family ranges (160 families for CounterFact and zsRE; 128
  for WikiData-CounterFact). Connecting lines only guide the eye across
  categorical endpoints. Because model, configuration, benchmark, and probe
  change together, the trajectories document heterogeneity rather than isolate
  its cause.}
  \label{fig:appendix-endpoint-trajectories}
\end{figure}

\input{sections/appendices/wikirecent_prospective_study.tex}
\input{sections/appendices/gpt2xl_endpoint_transport.tex}
\input{sections/appendices/llama_recurrence_test.tex}
\input{sections/appendices/end_to_end_and_multiplicity.tex}

\subsection{Summary of invalid shortcuts}

\begin{table}[!htbp]
  \centering
  \footnotesize
  \setlength{\tabcolsep}{3pt}
  \caption{\textbf{Four unsupported shortcuts and evidence-based alternatives.}}
  \label{tab:decision-replacements}
  \begin{tabularx}{\linewidth}{@{}>{\hsize=.78\hsize}Y
      >{\hsize=1.15\hsize}Y>{\hsize=1.07\hsize}Y@{}}
    \toprule
    Unsupported shortcut & Relevant evidence & Evaluation alternative \\
    \midrule
    Equal update counts imply the same pre-answer history
      & 24 schedule labels produce 1, 4, 12, or 24 history classes
      & Record the ordered updates before each answer \\
    Immediate learning implies durable, local learning
      & LoRA: acquisition passes 6/6; retention and locality pass 0/6
      & Test acquisition, retention, and locality separately \\
    A method ordering carries over to a new setting
      & Observed GPT-2XL and Llama-3.2-3B orderings differ across evaluated bundles; the cross-model mapping is incomplete
      & Evaluate the same methods directly in the new setting \\
    An average improvement implies usability
      & WikiRecent mean improvement passes while locality fails
      & Require every prespecified behavioral check to pass \\
    \bottomrule
  \end{tabularx}
\end{table}

\section{Reproducibility materials and scope}
\label{app:artifact}
\label{app:artifact-boundary-v93}

We plan a public release of the tracing code, aggregate results used in the
figures and tables, sanitized WikiRecent sufficient statistics, validation
scripts, and upstream notices. The release will omit legacy family-level
measurements, raw or restricted data, model checkpoints, private execution
records, randomized policy files, and third-party implementation details or
histories. The included WikiRecent statistics will support reanalysis of that
study; legacy family-level intervals cannot be independently recomputed from
the aggregate results alone.

\paragraph{Scope of the combined framework.}
The framework builds on intervention variants and exposure mappings to distinguish
an assigned plan from the updates actually completed; program slicing and
provenance to record operations that may affect an answer; model-editing metrics
to assess acquisition, retention, and locality; and transportability work to
separate evidence in the original and new settings. Our contribution is to
connect these existing ideas in one evaluation record. Tracing guarantees
depend on the declared score dependencies, and comparisons in an unevaluated
setting require additional assumptions or target evidence.

%% file: sections/appendices/evidence_contract.tex
\subsection{Evidence records and decision rules}
\label{sec:typed-evidence-record}

An evidence record contains fields for update histories and their resolution,
effect estimates and uncertainty, acquisition, retention, locality, and cost.
Effects compare policies; feasibility tests absolute requirements for each
policy. Comparisons across settings additionally specify the correspondence
between policies and the relation being tested. Missing measurements remain
explicit, and estimates from assays with different outcome meanings or units
are kept separate. We use \emph{context} and \emph{setting} synonymously;
$c$ denotes the setting defined in the main text, and $e_c$ denotes its endpoint.

Let $q_c$ be a nonzero, zero-sum contrast over the evaluated policy catalogue
$\Pi_c^{\rm eval}$. A named estimand $m$ specifies the intervention being
compared, outcome transform, aggregation, orientation, and endpoint;
$u_c^m$ is its inference unit. Its
local effect record is
\begin{equation}
\mathsf{Eff}_c^m(q_c)=
\big(m,\widehat\Delta_c^m(q_c),I_c^m(q_c),H_c^m(q_c),u_c^m,
      \mathsf{Asn}_c^m,\mathsf{Int}_c^m\big),
\end{equation}
where $I$ is the declared uncertainty interval, $H\in\mathcal S$ is the
protocol's effect decision,
$\mathsf{Asn}_c^m\in\{\textsc{prospective},\allowbreak\textsc{descriptive},
\textsc{diagnostic}\}$ records when the analysis was fixed, and
$\mathsf{Int}_c^m\in\{\textsc{valid},\textsc{invalid}\}$ records provenance and
integrity separately. Superscripts $Z$, $K$, $T$, and $\iota$, and the gate
subscript $j$, use these same timing and integrity vocabularies. Here
$\mathcal S=\{\textsc{pass},\allowbreak\textsc{fail},\allowbreak\textsc{inconclusive},
\textsc{undefined}\}$. Its conjunction operator $\bigwedge_{\mathcal S}$ uses
the precedence \textsc{fail} $>$ \textsc{undefined} $>$
\textsc{inconclusive} $>$ \textsc{pass}.
We write $\bot$ for a field that was not measured or not declared, and
$\mathcal M_c$ for the set of named estimands in context $c$.

Let $\mathcal D_{c,e_c}$ be the declared score-dependency adapter and
$\mathsf{Cert}_{c,e_c}$ its source/input binding. The resolution record
\begin{equation}
\begin{aligned}
\mathsf{Res}_{c,e_c}(q_c)=\big(&\mathcal D_{c,e_c},
\{w_r,z_{r,s,e_c}\}_{r,s},\mathsf{Alias}_{c,e_c},
\mathsf{Exposure}_{c,e_c},\\[-1mm]
&s_{Z,c,e_c}(q_c),S^Z_{c,e_c}(q_c),\mathsf{Asn}^Z_{c,e_c},
\mathsf{Int}^Z_{c,e_c},\mathsf{Cert}_{c,e_c}\big)
\end{aligned}
\end{equation}
retains the recorded traces, row weights, aliases, exposure,
resolution, protocol-specific support status $S^Z\in\mathcal S$, timing,
integrity, and certificate. A gate record is
$G_{c,j}(\pi)=(j,\allowbreak\psi_{c,j},\allowbreak\widehat\nu_{c,j}(\pi),\allowbreak I_{c,j}(\pi),
\allowbreak S_{c,j}(\pi),\allowbreak\mathsf{Asn}_{c,j}(\pi),\allowbreak\mathsf{Int}_{c,j}(\pi))$, containing
the gate identity, declared rule, estimate, interval, status
$S_{c,j}(\pi)\in\mathcal S$, analysis timing, and integrity. By
construction, $S_{c,j}=\textsc{undefined}$ when a required measurement or its
integrity record is missing or invalid; a valid but nonprospective gate is
\textsc{fail} whenever prospective timing is part of the declared rule. Retain
the gate records for each policy separately in
$\mathbf G_c(q_c)=(G_{c,j}(\pi))_{\pi\in\operatorname{supp}(q_c),j}$. A separate cost record
$K_c(\pi)=(\boldsymbol\kappa_c(\pi),\mathbf u_c^K,g_c,\mathcal B_c,
\mathsf{Asn}_c^K,\mathsf{Int}_c^K)$ reports named cost coordinates and units, an optional
prespecified scalarization $g_c$, a finite protocol budget or
$\mathcal B_c=\bot$, and cost-record integrity. When $\mathcal B_c<\infty$,
$g_c(\boldsymbol\kappa)$ must have the budget's units; when no scalar budget was
declared, $g_c=\mathcal B_c=\bot$. A vector such as outer-call counts is not
silently scalarized or compared across writers. Let
$\mathbf K_c(q_c)=(K_c(\pi))_{\pi\in\operatorname{supp}(q_c)}$. The local
evidence record is
\begin{equation}
\Gamma_c^m(q_c)=
\big(\mathsf{Res}_{c,e_c}(q_c),\mathsf{Eff}_c^m(q_c),\mathbf G_c(q_c),
\mathbf K_c(q_c)\big).
\label{eq:local-control-record}
\end{equation}
It exists without a target context and can therefore represent the paper's
resolution, effect, and feasibility studies on their own terms. For a
resolution-only study with no outcome measurement, set
$\widehat\Delta=I=\bot$ and $H=\textsc{undefined}$; $\bot$ records absence,
not a numerical zero. When several named estimands are reported, the empirical
record is the bundle $(\Gamma_c^m)_{m\in\mathcal M_c}$; it is not collapsed to
one scalar. Formally, the
base \emph{Control Map} is the partial function
\begin{equation}
\mathfrak C:(c,q_c,m_c)\ \rightharpoonup\ \Gamma_c^{m_c}(q_c),
\label{eq:control-map}
\end{equation}
whose domain requires a well-formed declared contrast, dependency adapter, and
valid trace provenance. Zero resolution remains in the domain. Under the
assumptions of Equation~\ref{eq:bounded-resolution-effect}, its positive and
negative trace mixtures coincide and the realized assignment contrast is zero;
individual policies need not be pairwise aliased. If the intended estimand
requires distinct implemented mixtures, its effect field is \textsc{undefined}
rather than a null estimate. Policy-level feasibility fields may still be evaluated.
Positive contrast support is required only before estimating the intended
effect.

Transport is an optional extension of two local records and requires an
explicit policy correspondence. Before target outcomes are opened, define
\begin{equation}
 \iota_{c\rightarrow c^*}:\operatorname{supp}(q_c)
 \longrightarrow \Pi_{c^*}^{\rm eval},
 \qquad q_{c^*}=\iota_{c\rightarrow c^*\#}q_c,
 \label{eq:policy-correspondence}
\end{equation}
where the pushforward sums source coefficients mapped to each target policy.
The correspondence is admissible only when $q_{c^*}$ remains nonzero and the
predeclared positive-versus-negative orientation of the contrast is preserved;
otherwise transport is \textsc{undefined}. Its certificate records which
model, writer, support, strength, execution template, and endpoint fields are
held fixed or changed. The policy correspondence record
$\mathsf{Iota}_{c\rightarrow c^*}=(\iota,\mathsf{Cert}^{\iota},
\mathsf{Asn}^{\iota},\mathsf{Int}^{\iota},S^{\iota})$ records whether the
mapping was fixed before target outcomes were accessed and whether its
provenance is valid.
If it is absent, the target contrast is not silently inferred from matching
names.

Define the target local record analogously and let
$\mathbf R_{c\rightarrow c^*}(q_c,q_{c^*})=
(s_{Z,c,e_c}(q_c),s_{Z,c^*,e_{c^*}}(q_{c^*}))$. With
$\mathbf m=(m_c,m_{c^*})$, let the transport record be
$\mathsf{Tr}_{c\rightarrow c^*}^{\mathbf m}=(\rho,T,
\mathsf{Asn}^T,\mathsf{Int}^T)$, where $\rho$ is the declared relational rule
and $T\in\mathcal S$ its result. The transport-extended record is
\begin{equation}
\begin{split}
\Gamma^{\mathbf m,\iota}_{c\rightarrow c^*}(q_c)=
\big(&\mathsf{Iota}_{c\rightarrow c^*},
\Gamma_c^{m_c}(q_c),\Gamma_{c^*}^{m_{c^*}}(q_{c^*}),
\mathbf R_{c\rightarrow c^*}(q_c,q_{c^*}),\\[-1mm]
&\mathsf{Tr}_{c\rightarrow c^*}^{\mathbf m}(q_c,q_{c^*})\big).
\end{split}
\label{eq:typed-control-record}
\end{equation}
The optional extension for transport is
\begin{equation}
\mathfrak C^{\rightarrow}:
(c,c^*,q_c,\mathsf{Iota},\mathbf m)
\ \rightharpoonup\
\Gamma^{\mathbf m,\iota}_{c\rightarrow c^*}(q_c).
\label{eq:transport-control-map}
\end{equation}
The extension requires well-formed source and target contexts with valid
provenance. The value $\mathsf{Iota}=\bot$ represents an attempted transport
claim with no supplied policy correspondence; in that case,
$S^\iota=T=\textsc{undefined}$. The transport field $T\in\mathcal S$ records
the result of the declared relation, such as preservation of effect direction
under the certified policy correspondence. If a required correspondence, target
trace, outcome, or instrument is absent, $T$ is \textsc{undefined}. A local
record does not require a target setting. Missing measurements remain explicit
rather than being imputed or interpreted as zero. Each field retains its role:
$\mathbf R$ records design geometry, each $\mathsf{Eff}$ records an estimate,
decision, and provenance, $\mathbf G$ records absolute gates for each policy,
$\mathbf K$ records costs and any declared budget, and $T$ records the relation
between settings. Adding or averaging these fields has no defined meaning.

Let $\mathcal R^{\rm pass}\subseteq(0,1]^2$ be the protocol-declared set of
source--target resolution pairs that license its transport analysis. For the
strongest prospective transport claim, let $S^\iota=\textsc{pass}$ when the
correspondence is present, admissible, prospectively fixed, and valid;
\textsc{fail} when it is valid but not prospectively fixed; and
\textsc{undefined} when it is absent, inadmissible, or invalid. Let
$S^R=\textsc{pass}$ when both resolution records are valid and prospective and
their positive pair lies in $\mathcal R^{\rm pass}$; \textsc{undefined} when
either record is unavailable, invalid, or has zero resolution; and
\textsc{fail} when the positive records are valid but retrospective or their
pair lies outside $\mathcal R^{\rm pass}$. Finally, let
$S_c^{P,m}=\textsc{pass}$ when $\mathsf{Int}_c^m=\textsc{valid}$ and
$\mathsf{Asn}_c^m=\textsc{prospective}$, \textsc{fail} for a valid but
nonprospective record, and \textsc{undefined} for invalid or missing
provenance. Define $S^{T,P}$ by preserving $T$ when its rule is valid and
prospectively fixed, setting it to \textsc{fail} when valid but nonprospective,
and to \textsc{undefined} when the rule or integrity record is absent or
invalid. Then define
\begin{equation}
\begin{aligned}
&\mathsf{TransportSupported}^{\mathbf m}_{c\rightarrow c^*}(q_c)\\[-1mm]
&\quad=\bigwedge_{\mathcal S}\!\left(
S^\iota,S^R,S_c^{P,m_c},S_{c^*}^{P,m_{c^*}},\right.\\[-1mm]
&\hspace{35mm}\left.
H_c^{m_c}(q_c),H_{c^*}^{m_{c^*}}(q_{c^*}),
S^{T,P}_{c\rightarrow c^*}(q_c,q_{c^*})\right).
\end{aligned}
\label{eq:evidence-supported}
\end{equation}
This predicate requires prospective evidence for the transport claim.
Descriptive and diagnostic records retain their separate timing labels.

For a policy $\pi^\dagger\in\Pi_{c^*}^{\rm eval}$ considered after target
evaluation, let $\mathcal J_{c^*}^{\rm req}$ index the protocol's required
behavioral gates. Let $S^K_{c^*}(\pi)=\textsc{pass}$ when
$\mathcal B_{c^*}=\bot$ (cost is then neutral, not certified) or when a valid,
prospectively fixed finite-budget record satisfies
$g_{c^*}(\boldsymbol\kappa_{c^*}(\pi))\leq\mathcal B_{c^*}$; let it be
\textsc{fail} for a valid prospective budget exceedance or a valid but
retrospectively selected budget, and \textsc{undefined} when a required cost
record is absent or invalid. Define absolute target feasibility by
\begin{equation}
\mathsf{TargetFeasible}_{c^*}(\pi^\dagger)
=\bigwedge_{\mathcal S}\!\left(
S^K_{c^*}(\pi^\dagger),
\{S_{c^*,j}(\pi^\dagger):j\in\mathcal J_{c^*}^{\rm req}\}\right).
\label{eq:target-deployable}
\end{equation}
Thus cost is always a reported coordinate but is a pass/fail requirement only
for a protocol that declares a finite budget and a matching scalar cost
coordinate or prespecified scalarization. A resolved failure suffices to reject
the conjunction; otherwise a missing required component propagates
\textsc{undefined}; otherwise any unresolved component propagates
\textsc{inconclusive}. The empirical ``full gate'' in this paper sets
$\mathcal B_c=\bot$ and conjoins acquisition, retention, and locality;
it must not be read as satisfying an unstated deployment budget.
For a pairwise target contrast, let
$\operatorname{cand}(q_{c^*})\in\operatorname{supp}(q_{c^*})$ be the arm
prospectively designated as positive/preferred; a more general contrast
requires an equally explicit designation rule. Let $S^{\rm cand}=\textsc{pass}$
when $\pi^\dagger$ is that designated candidate, \textsc{undefined} when no
designation exists, and \textsc{fail} when another policy is substituted. The
combined evidence requirement is
\begin{equation}
\begin{aligned}
&\mathsf{EvidenceComplete}^{\mathbf m}_{c\rightarrow c^*}
(q_c,\pi^\dagger)\\[-1mm]
&\quad=\bigwedge_{\mathcal S}\!\left(
S^{\rm cand},
\mathsf{TransportSupported}^{\mathbf m}_{c\rightarrow c^*}(q_c),\right.\\[-1mm]
&\hspace{35mm}\left.
\mathsf{TargetFeasible}_{c^*}(\pi^\dagger)\right).
\end{aligned}
\label{eq:claim-ladder}
\end{equation}
This decision applies to one target policy after evaluation.
Equation~\ref{eq:target-deployable} assesses $\pi^\dagger$'s absolute
feasibility; the record retains the comparator's gates separately. An
all-\textsc{pass} component vector can be summarized as \textsc{pass}.
Mixed \textsc{fail}, \textsc{inconclusive}, and \textsc{undefined} component
statuses remain explicit.

\paragraph{Evaluation procedure.}
The formalism induces a five-step evaluation procedure. (1) Freeze the
candidate catalogue, endpoint, contrasts, dependency adapter, gates, and
source-to-target correspondence. (2) Compile traces before opening outcomes
and mark zero-resolution contrasts unavailable. (3) Estimate effects only for
supported contrasts, retaining uncertainty and analysis timing. (4) Evaluate
acquisition, retention, and locality as separate absolute gates for each
candidate, record cost separately, and apply it as a gate only if a budget was
declared. (5) Test the frozen relational transport rule in the target setting
and otherwise return the unresolved or undefined component rather than a
winner. This procedure produces an evidence record containing histories and
their resolution, effect estimates and uncertainty, absolute gates, costs,
and any transport result. The experiments below measure different components
of this record in controlled settings.

\paragraph{What each component establishes.}
Positive resolution means that the recorded histories support the contrast
for estimation. It does not establish causal identification or guarantee a
nonzero effect: Equation~\ref{eq:bounded-resolution-effect} permits a zero
effect at any positive resolution. Acquisition does not establish feasibility:
the tested LoRA writer acquires its targets but does not pass the joint
retention and locality gate. Feasibility in one setting does not establish a
stable comparison in another: fixed MEMIT passes the named CounterFact
criterion, but its locality advantage over ROME reverses at both tested
KnowEdit endpoints. These observations distinguish resolution $\nRightarrow$
effect, effect $\nRightarrow$ feasibility, and source feasibility
$\nRightarrow$ cross-endpoint verdict stability. The evidence record therefore
reports resolution, effects, absolute feasibility, and transport separately.

%% file: sections/appendices/trace_resolution_theory.tex
\subsection{Trace-resolution consequences}
\label{app:resolution-calculus}

Definition~\ref{def:trace-resolution} gives the deterministic and stochastic
history laws, fixes the total-variation convention, and states the exact
score-sufficiency premise. The deterministic formula is derived in
Appendix~\ref{app:proof-resolution-formula}, and the complete Markov-contraction
and bounded-score argument is in
Appendix~\ref{app:proof-trace-resolution-certificate}.

At $s_Z=0$, the positive and negative implemented trace mixtures coincide, but
individual policies need not be pairwise aliased. Their assignment contrast is
zero. If the named estimand requires distinct implemented mixtures, the
record marks that intended scientific contrast \textsc{undefined}; it does not
reinterpret the zero as an estimate of distinct treatment versions.
The bound does not apply if an
action changes the row distribution or an omitted state path reaches the
score; those are failures of the declared sufficiency premise.
No finite magnitude cap follows for unbounded log probability or negative log
likelihood without clipping, a bounded transform, or additional tail
conditions; the empirical cap used in the paper is for binary correctness.

For any measurable summary $h(R,Z)$, define
\begin{equation}
s_h(q)=\operatorname{TV}\!\left(
h_{\#}P_{R,Z}^{q^+},h_{\#}P_{R,Z}^{q^-}\right).
\label{eq:summary-contraction}
\end{equation}
Theorem~\ref{thm:trace-resolution-certificate} gives
$0\leq s_h(q)\leq s_Z(q)$. Equality holds when a Hahn
sign partition for the two contrast measures is measurable through $h$; in
particular, it holds when the full pair $(R,Z)$ is recoverable almost surely
from $h$. Recovering $Z$ alone need not suffice because mixing rows can cancel
contrast mass. Strict inequality is information discarded by the summary.
Strict contraction can occur even when the observed task metric is unchanged.

\subsection{Why counts can fail, and when they are sufficient}
\label{app:count-sufficiency}

Fix a context $c$ and let every admissible update atom $a$ induce a
deterministic map $T_a^c$ on the complete mutable state. For an ordered word
$\Gamma=a_1\cdots a_T$, define
$R_c(\Gamma)=T_{a_T}^c\circ\cdots\circ T_{a_1}^c$ and let $\nu(\Gamma)$ count
each atom.

We use the standard commutativity characterization in
Lemma~\ref{lem:count-sufficiency} to identify when per-atom count vectors
determine the final-state map over all finite words. The proof is in
Appendix~\ref{app:proof-count-sufficiency}. On a restricted catalogue,
factorization instead requires that admitted words with the same count vector
induce the same map; pairwise commutativity remains sufficient but need not be
necessary. Scalar write counts discard the atom identities retained by $\nu$,
and an intermediate read can depend on the update prefix even when the final
maps commute.

\paragraph{Compiler soundness under declared assumptions.}
Given a common initial state and randomness, a complete adapter-relative trace,
and represented state sufficient for the score, equal traces induce the same
score law. Unequal traces permit but do not prove an effect. Given policies
$\Pi$, rows $\mathcal R$, endpoint $e$, dependency map $\mathcal D$, and
contrast $q$, the compiler returns
\begin{equation}
\mathsf{Compile}(\Pi,\mathcal R,e,\mathcal D,q)\mapsto
\big(\{z_{r,s,e}\},\{D_r\},\mathsf{alias},\mathsf{exposure},s_Z(q),
\mathsf{certificate}\big).
\label{eq:trace-compiler}
\end{equation}
It replays each policy through the endpoint, retains the transitions specified
by the dependency map, canonicalizes the ordered trace, forms rowwise
equivalence classes, and computes resolution. Before outcomes are opened, it
emits a certificate linking the source code and inputs to the resulting
traces. The compiler determines which comparisons have design support;
outcomes determine their sign and magnitude.

\subsection{Why separate coordinate studies do not identify a joint controller}
\label{app:controller-identification}

Let $\mathcal C$ and $\mathsf E$ denote the context and endpoint sets, and let
$\mathcal H\subsetneq\mathcal C\times\Pi\times\mathsf E$ be the tested
context--policy--endpoint tuples and let $x^*\notin\mathcal H$. Over the
unrestricted class of bounded response surfaces, any two distinct values in
the outcome range can be assigned at $x^*$ by two surfaces that agree at every
point in $\mathcal H$. Hence separately measured coordinate effects do not
identify a composed controller outside $\mathcal H$ without a declared
separability, smoothness, or invariance restriction.

For a concrete two-coordinate case, suppose the design observes
$Y_{00},Y_{10},Y_{01}$ but not $Y_{11}$, with $Y\in[a,b]$. Then
\begin{equation}
Y_{11}-Y_{10}-Y_{01}+Y_{00}
\in[a-Y_{10}-Y_{01}+Y_{00},\ b-Y_{10}-Y_{01}+Y_{00}].
\label{eq:missing-crossed-cell}
\end{equation}
The missing cell may take any value in $[a,b]$; separate coordinate effects do
not determine their composition.

If a prospectively fixed, commensurable utility $U_c$ compares two policies
$\pi_a,\pi_b$ feasible in contexts $c_1,c_2$ and
\begin{equation}
[U_{c_1}(\pi_a)-U_{c_1}(\pi_b)]
[U_{c_2}(\pi_a)-U_{c_2}(\pi_b)]<0,
\label{eq:ranking-reversal}
\end{equation}
then neither constant choice is optimal in both contexts within
$\{\pi_a,\pi_b\}$; a maximizing rule on this restricted set must distinguish
contexts. Abstention is a separate
option only after its utility is defined. We do not apply this
statement to endpoint-native metrics whose scales are not commensurable.

For a declared endpoint set $E\subseteq\mathsf E$, let $m_e$ be endpoint $e$'s
metric and let $I_e$ be the declared decision range for its estimated
within-endpoint contrast. Let $\rho_e=+$ when $\inf I_e>0$, $\rho_e=-$ when
$\sup I_e<0$, and $\rho_e=?$ otherwise. With
$\mathbf m=(m_e)_{e\in E}$, define the contrast and resolved-sign status
signature by
\begin{equation}
\widehat\Delta_e^{m_e}(\pi,\pi')=\widehat\mu_e^{m_e}(\pi)-\widehat\mu_e^{m_e}(\pi'),\qquad
\boldsymbol\sigma_E^{\mathbf m}(\pi,\pi')=
(\rho_e)_{e\in E}\in\{+,-,?\}^{|E|}.
\label{eq:transport-signature}
\end{equation}
When instruments differ, this is an endpoint-status signature rather than
evidence about one shared latent locality quantity or an identified causal
effect of dataset identity.

%% file: sections/appendices/adapter_robustness_and_gate_sensitivity.tex
\subsection{Adapter robustness, coverage evidence, and gate sensitivity}
\label{app:v95-robustness}

This section separates three questions that a trace audit can otherwise blur:
whether recorded executions conform to a declared trace, whether that trace
contains every score ancestor, and how conclusions change when either the
conditional-sufficiency premise or a gate threshold is varied. The sensitivity
calculations use a finite state machine and recorded results. The separate live
neural check measures conformance; editor ancestry-error rates remain unmeasured.

\paragraph{Conditional-kernel mismatch bound.}
The complete proof of Proposition~\ref{prop:kernel-mismatch}, including the
kernel-averaging and bounded-expectation steps, appears in
Appendix~\ref{app:proof-kernel-mismatch}. Here
$\epsilon_++\epsilon_-$ is one aggregate deviation budget; if instead each arm
is known only to be within $\epsilon$ of the common reference kernel $L$, the
bound uses $s_Z(q)+2\epsilon$ inside the cap at one.

\paragraph{Controlled hidden-cache state-machine ablation.}
Let $N\geq1$ and $m,k\geq0$ be integers with $m+k\leq N$, let $R$ be uniform
over the $N$ rows, and choose disjoint row sets $C,H$ of sizes $m,k$. The
positive arm records visible bit $V=\ind\{R\in C\}$ and writes hidden cache bit
$U=\ind\{R\in H\}$; the negative arm writes neither. The declared trace records
$V$ but omits $U$. Holding these executions and declared traces fixed, toggle
only the scorer: $Y_+^{(\lambda)}=\ind\{R\in C\}+\lambda\ind\{R\in H\}$,
$Y_-^{(\lambda)}=0$, for $\lambda\in\{0,1\}$. When $\lambda=0$, the cache is
non-ancestral and the declared exact bound holds. When $\lambda=1$, the cache
becomes an omitted score ancestor. For this mode, define the common reference
kernel on the declared input $x=(r,v)$ by
\[
L(\mathrm dy\mid r,v)=\delta_{v+\ind\{r\in H\}}(\mathrm dy).
\]
Because $C\cap H=\varnothing$, this is a binary-output kernel on every input
in the support of either arm. It matches the positive-arm kernel everywhere;
on the negative-arm support it differs from the constant-zero kernel exactly
on rows in $H$. Consequently,
\begin{equation}
s_Z(q)=\frac mN,\qquad \epsilon_+=0,\qquad
\epsilon_- = \frac kN,\qquad
\TV(P_Y^{q^+},P_Y^{q^-})=\frac{m+k}{N}.
\label{eq:v95-tightness-witness}
\end{equation}

\begin{proof}[Verification of Equation~\ref{eq:v95-tightness-witness}]
For each $r\in C$, the positive arm has declared input $(r,1)$ and the
negative arm has $(r,0)$; for each $r\notin C$, both arms have $(r,0)$.
Because the row coordinate is part of the input and every row has mass $1/N$,
the two declared-input laws differ on exactly the $m$ rows in $C$. Hence
$s_Z(q)=m/N$.

When $\lambda=0$, the common kernel
$L_0(\mathrm dy\mid r,v)=\delta_v(\mathrm dy)$ agrees with both arms on their
respective supports. Theorem~\ref{thm:trace-resolution-certificate} therefore
applies exactly, and the output total variation is $m/N=s_Z(q)$.

Under the positive arm, $v=\ind\{r\in C\}$, so the reference kernel returns
$v+\ind\{r\in H\}=Y_+^{(1)}$ on every row. Thus $\epsilon_+=0$. Under the
negative arm, $v=0$ and the actual output is zero, whereas the reference
kernel returns one exactly for $r\in H$. The pointwise total variation is
therefore $\ind\{r\in H\}$, whose uniform-row average is
$\epsilon_-=k/N$.

Finally, $P_Y^{q^+}$ is Bernoulli with success probability $(m+k)/N$, while
$P_Y^{q^-}=\delta_0$. Their total variation is the positive law's mass at one,
namely $(m+k)/N$. This equals $s_Z(q)+\epsilon_++\epsilon_-$, proving that the
bound in Proposition~\ref{prop:kernel-mismatch} is attained. If the declared
history is expanded to $(V,U)$, the two arms differ exactly on $C\cup H$, so
$s_Z^*=(m+k)/N$; the common kernel $\delta_{v+u}$ then restores the premise and
bound of Theorem~\ref{thm:trace-resolution-certificate}.
\end{proof}

Thus the incomplete exact bound misses by exactly $k/N$,
Proposition~\ref{prop:kernel-mismatch} is tight, and expanding the trace to
include $U$ restores Theorem~\ref{thm:trace-resolution-certificate} with
$s_Z^*=(m+k)/N$. A CPU check evaluates both scorer modes on all 16
cells with $N=1000$, $m\in\{0,1,10,100\}$, and
$k\in\{0,1,10,50\}$. It verifies the non-ancestral bound in 16/16 cells, the
omitted-ancestor failure in all 12 cells with $k>0$, and restored coverage in
16/16 cells. These results concern the finite state machine defined above.

\paragraph{Live neural conformance check.}
The CPU protocol was recorded locally before its randomly initialized model
was constructed, without external preregistration. Its design was informed by
two earlier pilots. It instantiates a two-layer Llama with PEFT-LoRA on the query and
value projections. Every one of 16 frozen rows performs exactly two SGD writes
in A-then-B order and reaches one common terminal model and optimizer state;
only whether B precedes the score changes. The negative mixture contains two
full pre-score traces and six A-only pre-score traces, while the positive
mixture contains six and two, respectively, giving $s_Z=1/2$.

The runtime records trainable-parameter transitions and nonzero scalar-score
gradient paths, together with the state needed to check execution continuity.
The compiler validates all 16 chronological chains, rejects 16 corruptions
specified in advance (ten fault families plus six compiler-closure faults),
and accepts the restored trace. The check establishes conformance of the
recorded paths in one randomly initialized model. Coverage of unrecorded or
zero-gradient score dependencies remains unverified.

For the SocialIQA swaps, $s_Z=1/1954$. Table~\ref{tab:v95-social-budgets}
shows how a declared aggregate budget
$\bar\epsilon=\epsilon_++\epsilon_-$ changes the binary-outcome cap. These are
conditional calculations: the study does not supply an empirical value of
$\bar\epsilon$.

\begin{table}[H]
\centering
\caption{\textbf{SocialIQA certificate sensitivity.} The right column is
$100\min\{1,1/1954+\bar\epsilon\}$ percentage points.}
\label{tab:v95-social-budgets}
\small
\begin{tabular}{rr}
\toprule
Declared aggregate deviation $\bar\epsilon$ & Binary-effect cap (points) \\
\midrule
$0$ & $0.0512$ \\
$0.001$ & $0.1512$ \\
$0.010$ & $1.0512$ \\
$0.050$ & $5.0512$ \\
\bottomrule
\end{tabular}
\end{table}

\paragraph{Conformance is not ancestry coverage.}
Table~\ref{tab:v95-adapter-assurance} reports the strongest claim supported at
each recorded boundary. A conformance check asks whether an observed or
replayed execution matches the declared trace. A coverage check would require
an independent oracle for the complete set of score ancestors. No such oracle
was recorded for EasyEdit internals, so false-negative and false-positive
ancestry rates are unavailable rather than zero. A false negative can violate
the common-kernel premise. A false positive does not omit a score path, but can
split outcome-equivalent policies and make the quotient unnecessarily fine.

\begin{table}[H]
\centering
\caption{\textbf{Conformance and coverage.} Counts establish
conformance only at the level named in the second column.}
\label{tab:v95-adapter-assurance}
\scriptsize
\begin{tabularx}{\textwidth}{p{0.17\textwidth}p{0.23\textwidth}p{0.29\textwidth}Y}
\toprule
Study & Recorded boundary & Conformance evidence & Ancestry-coverage claim \\
\midrule
MAC trace atlas & Declared deterministic scheduler graph & Complete 27,072-row enumeration; 184 outcome-blind checks & Exact relative to the declared scheduler adapter; no whole-program claim \\
Constructed neural LoRA test & Recorded live state transitions and nonzero score-gradient paths & 16/16 chains validate; $s_Z=1/2$; 16 checks reject and restoration passes & Conformance for recorded paths in one randomly initialized small Llama; unrecorded dependencies remain unverified \\
GPT-2XL LoRA phase study & Declared PEFT-LoRA/SGD event and optimizer-step trace & Separate implementation matches 3,072/3,072 fields; 15/15 tests reject five omission/alias/substitution fault classes & Structural reconstruction only; no live-state observation or completeness theorem \\
WikiRecent & Score-ancestral editor outer calls & 512/512 policy rows match precomputed certificates & EasyEdit-internal false-positive/negative rates unavailable \\
GPT-2XL endpoints & Reconstructed editor outer-call traces & 448/448 compiled cells agree; 1,792 family--order units and 3,584 policy traces match & Editor internals were not observed \\
Llama-3.2-3B endpoints & Declared dose-four/terminal atomic outer-call traces & Post-outcome projection: 3,584/3,584 rows conform; all 1,792 paired traces separate ($s_Z=1$) & EasyEdit internals and cross-bundle correspondence unavailable; diagnostic component only \\
\bottomrule
\end{tabularx}
\end{table}

\paragraph{Exact locality-threshold breakpoints.}
Let $d\geq0$ be an allowed damage in percentage points, so the success
threshold is $-d$. For point $\widehat L$ and range $[\ell,u]$ with
$\ell\leq\widehat L\leq u\leq0$, the main-text joint rule is
\textsc{fail} when $d<-u$, \textsc{inconclusive} when
$-u\leq d\leq-\ell$, and \textsc{pass} when $d>-\ell$.
Under this containment assumption, the joint rule reduces to the interval
rule, so $\widehat L$ introduces no additional breakpoint.
Table~\ref{tab:v95-threshold-breakpoints} applies this rule to four frozen
records and preserves their pass or nonpass decisions at five points.

\begin{table}[H]
\centering
\caption{\textbf{Locality verdict as a function of damage tolerance $d$.}
Endpoint records are not pooled. Bounds are percentage points.}
\label{tab:v95-threshold-breakpoints}
\scriptsize
\begin{tabularx}{\textwidth}{p{0.23\textwidth}p{0.15\textwidth}YYYc}
\toprule
Context & Policy & \textsc{fail} & \textsc{inconclusive} & \textsc{pass} & At $d=5$ \\
\midrule
New 160-family CounterFact panel & MEMIT-high & $d<1.125000$ & $1.125000\leq d\leq3.203125$ & $d>3.203125$ & \textsc{pass} \\
New 160-family CounterFact panel & ROME-low & $d<3.078125$ & $3.078125\leq d\leq6.421875$ & $d>6.421875$ & \textsc{inconclusive} \\
WikiRecent shared locality & MEMIT-high & $d<33.984375$ & $33.984375\leq d\leq52.734375$ & $d>52.734375$ & \textsc{fail} \\
WikiRecent shared locality & ROME-low & $d<32.2265625$ & $32.2265625\leq d\leq50.781250$ & $d>50.781250$ & \textsc{fail} \\
\bottomrule
\end{tabularx}
\end{table}

The six selected LoRA settings have point damage from $30.625$ to $53.203$
points. Retention passes in 0/6, so relaxing locality alone cannot produce a
full pass. Under the main-text joint rule, equality at either interval boundary
is \textsc{inconclusive}. A rule that gives failure precedence to a point miss
would instead label some threshold-crossing ROME results as \textsc{fail}.
Every observed WikiRecent interval lies wholly below the five-point threshold,
so both conventions give the same four WikiRecent verdicts and headline results.

The prospective WikiRecent record also exposes the closest non-locality
threshold: ROME's rephrase installation point is $0.20703125$, only
$0.00703125$ above the frozen $0.20$ cutoff (its descriptive reweighting range
crosses that cutoff, but the declared installation gate is point-based).
Raising the installation cutoff above $0.20703125$ would change that component
to \textsc{fail}; it would not change the existing full verdict, which already
fails locality. All four WikiRecent acquisition and continuous-retention ranges
clear zero, and every retained-among-installs point clears $0.50$.

\paragraph{Controlled locality-readout sensitivity.}
All three summaries use the same 1,280 editor-by-order CounterFact rows, reset
and terminal checkpoints, ten neighborhood prompts per family, family weights,
and 10,000 paired-bootstrap index draws; only the readout changes.
MEMIT-minus-ROME is $+2.562500$ percentage points
$[+1.062500,+4.265625]$ for neighborhood success, where higher is better;
$-0.094463$ $[-0.151609,-0.044748]$ for full-vocabulary KL from reset; and
$-0.111094$ $[-0.135156,-0.088750]$ for greedy-next-token change, where lower
is better for the latter two. The primary success endpoint and secondary
diagnostics were encoded in the frozen analysis; interpreting their agreement
as factor-isolation evidence is post hoc. This fixes neither probe selection
nor cross-bundle factors, so it rules out dependence on the binary score at
this source point but does not attribute the observed reversals. Relative
ordering is robust across the three readouts; absolute feasibility under KL,
greedy change, or UniEdit remains \textsc{undefined} without separately frozen
gates.

Threshold sensitivity and instrument sensitivity are different audits. The
table varies $d$ while holding the locality probe and estimator fixed. Replacing
binary relation-specificity with full-vocabulary KL, greedy token change, or a
compositional benchmark instead defines a new metric $m$, gate, or target
context and requires a separately frozen record. The post-hoc WikiRecent KL and
token-change diagnostics point in the same damage direction but reuse the same
two probes, so they add metric breadth rather than probe breadth. Gate values
should therefore be tied to an explicit deployment loss or tolerance and frozen
per endpoint; the Control Map does not nominate five points as a universal
calibration.

%% file: sections/appendices/matched_factual_recall_stage0.tex
\subsection{Matched factual-recall Stage-0 protocol and full decision}
\label{app:stage0-factual}

\paragraph{Protocol and study design.}
The Stage-0 protocol was fixed before validation-run or study outcomes. It crosses four
fixed settings (Pythia-410M current, Pythia-410M step100000, GPT-2 Large, and
GPT-2XL), four LoRA initializations, three doses $\{1,2,4\}$, and four
eight-family shards. The resulting 192 evaluation runs cover 32 CounterFact families and
produce 1,536 rows over family, model, initialization, and dose. Each row includes a direct
prompt, two unseen paraphrases, terminal recall after all four writes, and ten
unrelated-neighborhood prompts. At $d=4$ the query is terminal, so continuous
acquisition and retention coincide; $d=1,2$ include later writes. Timed and deferred policies execute
the same four actions and 16 optimizer steps and must end in identical
mutable state.

Before reviewing outcomes, we checked that all 192 planned runs were unique
and complete, with no retries. A metadata correction made before outcome
access preserved the experimental settings, outputs, and estimator.

\paragraph{Frozen rule.}
The family is the inferential unit; the four adapter initializations and two
paraphrases are averaged within family, followed by 10,000 deterministic
family-cluster bootstrap draws.  At the pre-outcome-frozen predicted dose,
acquisition requires a positive lower endpoint of the frozen 95\%
family-reweighting range for the mean
two-paraphrase timed-minus-deferred target-log-probability contrast.  Retention
requires a positive lower endpoint for terminal-minus-reset target log
probability, joint post-focal installation of at least $20\%$, and terminal
retention among installed pairs of at least $50\%$.  Locality requires a point
estimate of at least $-5$ percentage points and a lower endpoint above $-5$
for terminal-minus-reset neighborhood success. Stage 1 proceeds only if all
three clauses pass in at least two of four settings.

\begin{table}[H]
\centering
\caption{Full Stage-0 decision at each pre-outcome-frozen predicted dose.
``Call'' is the frozen selected dose, ``Peak'' is the empirical maximizing
dose, and ``Keep$\mid$install'' is retention among installed pairs.
``Acq.'' and ``Ret.'' denote acquisition and retention. Log-probability
contrasts are in nats; locality is in percentage points.}
\label{tab:stage0-full-decision}
\scriptsize
\resizebox{\textwidth}{!}{%
\begin{tabular}{lrrrrrrr}
\toprule
Setting & Call & Peak & Acq. [95\% range] & Ret. [95\% range] & Install & Keep$\mid$install & Locality [95\% range] \\
\midrule
Pythia current & 1 & 1 & $+8.481$ $[+7.531,+9.473]$ & $-10.947$ $[-13.035,-8.853]$ & $0.8125$ & $0.0096$ & $-37.03$ $[-49.61,-24.53]$ \\
Pythia step100000 & 1 & 2 & $+6.666$ $[+5.781,+7.572]$ & $-12.029$ $[-13.529,-10.609]$ & $0.3906$ & $0.0000$ & $-38.44$ $[-48.83,-27.42]$ \\
GPT-2 Large & 4 & 4 & $+4.463$ $[+3.841,+5.123]$ & $+4.463$ $[+3.841,+5.123]$ & $0.0781$ & $0.9000$ & $-53.20$ $[-62.81,-43.20]$ \\
GPT-2XL & 4 & 4 & $+2.746$ $[+2.041,+3.477]$ & $+2.746$ $[+2.041,+3.477]$ & $0.0547$ & $0.7143$ & $-41.95$ $[-51.88,-32.03]$ \\
\bottomrule
\end{tabular}}
\end{table}

\begin{table}[H]
\centering
\caption{Mean acquisition contrast across the complete Stage-0 dose grid.
These are descriptive point means; the frozen decisions use the predicted-dose
family-bootstrap ranges in Table~\ref{tab:stage0-full-decision}.}
\label{tab:stage0-dose-grid}
\small
\begin{tabular}{lrrr}
\toprule
Setting & $d_1$ & $d_2$ & $d_4$ \\
\midrule
Pythia current & $+8.481$ & $+3.225$ & $-6.114$ \\
Pythia step100000 & $+6.666$ & $+6.674$ & $-7.497$ \\
GPT-2 Large & $+0.520$ & $+1.340$ & $+4.463$ \\
GPT-2XL & $+0.342$ & $+1.101$ & $+2.746$ \\
\bottomrule
\end{tabular}
\end{table}

The acquisition clause passes 4/4 and the frozen peak calls match 3/4
numerical maxima.  Nevertheless, retention fails 4/4, locality fails 4/4, and
zero settings pass the conjunction. We therefore did not run Stage 1. This test
supports setting-conditioned acquisition and localizes distinct write-quality
failures, but it does not establish a safe writer, a population-level predictor
accuracy, or an online controller.

%% file: sections/appendices/factual_recall_family_extension.tex
\subsection{Natural factual-recall family extension}
\label{app:factual-family-extension}

The pre-outcome-frozen extension applies the Stage-0 estimator and gate to
  Qwen2.5-0.5B~\citep{qwen2024qwen25} and
  Llama-3.2-1B~\citep{meta2024llama32}. It contains 96 unique first-attempt runs and 768
sufficient-statistic rows: two models, four LoRA initializations, three
doses, and 32 CounterFact families. We checked completion before collecting
outcomes and independently recomputed the decision. Before reviewing any
outcomes, we corrected a collection check while preserving the evaluation
grid, row validation, and decision rule.

\begin{table}[H]
\centering
\caption{Family-extension acquisition surfaces and metrics at the frozen
dose-1 call. $A$ is acquisition, retention is a log-probability contrast in
nats, and $L$ is terminal-minus-reset neighborhood success in percentage points.}
\small
\begin{tabular}{lrrrr}
\toprule
Model & $A(d_1)$ [95\% range] & $A(d_2)$ & $A(d_4)$ & Retention (nats) / $L$ (pp) \\
\midrule
Qwen2.5-0.5B & $6.145$ $[4.737,7.556]$ & $-2.212$ & $-4.366$ & $-7.029$ / $-30.625$ \\
Llama-3.2-1B & $4.581$ $[3.837,5.358]$ & $0.333$ & $-1.671$ & $-4.143$ / $-38.359$ \\
\bottomrule
\end{tabular}
\label{tab:factual-family-extension}
\end{table}

Both frozen dose-1 calls equal the empirical finite-grid peak and both
acquisition gates pass.  Neither retention nor locality passes: retention
ranges are $[-8.104,-5.961]$ and $[-5.504,-2.833]$, locality ranges are
$[-41.328,-18.984]$ and $[-49.922,-25.938]$ percentage points, and retained
fractions among installed facts are zero.  Combining these two settings with
the prior four yields 5/6 exact calls for the frozen dose-selection rule versus
3/6 for always-dose-1 (two wins, four ties, zero losses). At $n=6$ this is a
descriptive selected-setting tally with no significance or population claim.

%% file: sections/appendices/editor_schedule_grids.tex
\subsection{Partially outcome-blind published-editor schedule grids}
\label{app:published-editor-dose-grids}

\paragraph{Evidence role.}
The analyses below reuse completed GPT-2XL Stage-0 outputs on the same 32
CounterFact families, four order replicates, two held-out paraphrases, terminal
retention endpoint, ten neighborhood prompts, and original gate definitions.
The analysis rules were fixed after the $d=4$ outcomes were known but
before the completed $d=1,2$ outputs were opened.  Thus they are partially
outcome-blind extensions, not fully prospective peak tests.  Empirical peaks
and winner--runner-up contrasts are descriptive. Dose 8 is
absent because the inherited Stage-0 family contains only four events.
Acquisition is evaluated at the dose query, while retention and locality are
evaluated at terminal after all four writes.  Dose-indexed Latin-square cells
can have different complete event orders and focal ranks. At $d=4$ the query is
terminal and continuous acquisition equals retention; $d=1,2$ include later writes.

\begin{table}[H]
\centering
\caption{GPT-2XL published-editor schedule grids. $A$ is acquisition at the
dose query in nats and $D$ is terminal locality damage after all four writes
in percentage points (lower is safer). Each entry is $A/D$. Native strength
values are not commensurate across editors.}
\label{tab:v81-editor-schedule-grids}
\small
\begin{tabular}{llrrrl}
\toprule
Editor & Native strength & $d=1$ & $d=2$ & $d=4$ & Frozen $A$-response label \\
\midrule
MEMIT & high & $7.185/1.875$ & $7.196/1.875$ & $7.205/1.875$ & \textsc{dose-insensitive} \\
ROME & low & $6.376/5.391$ & $6.513/5.469$ & $6.854/6.328$ & \textsc{inconclusive} \\
ROME & default & $8.118/18.281$ & $8.167/20.312$ & $8.201/25.547$ & \textsc{dose-insensitive} \\
ROME & high & $8.611/35.391$ & $8.625/38.906$ & $8.582/41.797$ & \textsc{dose-insensitive} \\
\bottomrule
\end{tabular}
\end{table}

For MEMIT-high, the max--min acquisition range is 0.019725 nats with a
paired-family 95\% empirical reweighting range $[0.004358,0.040115]$, wholly below the
frozen 0.5-nat practical-sensitivity threshold.  Joint installation is
$0.375/0.375/0.367$ and keep among installs is 1.0 at every dose; acquisition,
retention, locality, and the full gate pass at all three doses.  The nominal
$d=4$ minus $d=2$ lead is only 0.008073 nats (95\% bootstrap range
$[-0.002786,0.020460]$). This is classified \textsc{dose-insensitive}; it
does not turn the known $d=4$ maximum into a prospective prediction.

For ROME, the acquisition max--min ranges are 0.478 nats
($[0.210,0.764]$) at low strength, 0.083 ($[0.020,0.325]$) at default, and
0.043 ($[0.020,0.358]$) at high.  The frozen labels are respectively
\textsc{inconclusive}, \textsc{dose-insensitive}, and
\textsc{dose-insensitive}. All nine cells pass acquisition and retention,
but none certifies locality or the full conjunction. The low-strength locality-
response range is below the frozen 5-point dose-sensitivity margin; uncertainty
bands around the default/high 7.266/6.406-point ranges cross that margin, so
those response classifications are inconclusive.

The combined pattern is not evidence that dose and writer quality are
independent, nor that dose sensitivity diagnoses a weak writer: unsafe ROME
settings can be acquisition-flat. Query-time acquisition and terminal
quality both depend on the tested schedule/editor cell, but they are not a
same-terminal-state comparison. The LoRA terminal-locality mixtures differ across
dose labels, MEMIT-high is low-damage across its mixtures, and ROME's absolute
locality depends strongly on native strength even where query-time acquisition
is dose-insensitive.  The disjoint
64-family MEMIT--ROME primary comparison remains \textsc{inconclusive}, so
these discovery-panel grids neither establish universal safety nor rank the
methods causally.

We compared individual rows while holding editor, strength, the recorded
initialization or order replicate, family, and complete ordered case IDs
fixed. Terminal model state and locality matched exactly across all 160
MEMIT and 480 ROME cross-dose pairs. Together with 960 LoRA pairs, the comparison covers
1,600/1,600 pairs with zero mismatches.  Therefore movement of terminal
locality across dose-indexed cells is an order-mixture association, not an
identified effect of moving the read-only query boundary.

All 384 MEMIT rows and 1,152 ROME rows completed successfully on the first
attempt. An independent reanalysis reproduced both decisions and their
numerical summaries.

%% file: sections/appendices/memit_eight_event_study.tex
\subsection{Prospective shared-eight-event MEMIT facet}
\label{app:memit-eight-event}

Dose 8 is undefined in the inherited four-event Stage-0 family.  We therefore
froze a separate structural extension before executing any of its outcomes:
adjacent CounterFact families are paired into 16 disjoint blocks, both focal
orientations are retained, and each focal is followed by the other seven
distinct events in an order that varies across replicates. All four doses share the
identical eight-event sequence.  The inference unit is the disjoint family-pair
block; two focal orientations and four order replicates are averaged within
block before 10,000 paired block-bootstrap draws.

\begin{table}[H]
\centering
\caption{MEMIT-high on the separate shared-eight-event GPT-2XL facet.  $A$ is
query-time acquisition in nats and $D$ is terminal locality damage in
percentage points.  Brackets are pointwise 95\% block-reweighting ranges;
the max--min range and dose-eight-minus-dose-four comparison below use the
frozen max-deviation band.}
\label{tab:v82-memit-eight-event}
\small
\begin{tabular}{lrrrrr}
\toprule
Dose & $A$ [95\% range] & $D$ & Install & Retain/install & Full gate \\
\midrule
$d=1$ & $7.18483\ [6.21130,8.08745]$ & 1.09375 & 0.375 & 1.000 & \textsc{pass} \\
$d=2$ & $7.19700\ [6.22382,8.09705]$ & 1.09375 & 0.375 & 1.000 & \textsc{pass} \\
$d=4$ & $7.21356\ [6.25123,8.10839]$ & 1.09375 & 0.375 & 1.000 & \textsc{pass} \\
$d=8$ & $7.26836\ [6.30097,8.17701]$ & 1.09375 & 0.375 & 1.000 & \textsc{pass} \\
\bottomrule
\end{tabular}
\end{table}

The empirical maximum is dose 8.  Its difference from dose 4 is
$+0.05481$ nats with a max-deviation 95\% range $[+0.00180,+0.10781]$.
Thus the range remains above zero under the frozen joint-reweighting
rule, but the estimate is small relative to the prespecified 0.5-nat practical threshold.
The complete max--min range is 0.08353 nats with max-deviation range
$[0.03053,0.13653]$, yielding \textsc{dose-insensitive}. Terminal locality
damage is 1.09375 points and terminal-minus-reset retention is 7.26836 nats at
all four doses; every acquisition, retention, locality, and full gate passes.

Checks confirmed the same complete update order and terminal parameters for
every order replicate and focal orientation. Recorded reset, post-focal, and
terminal measurements, retention, locality, and recall also agreed across
doses. Only the read-only query checkpoint varied. Thus the small acquisition
contrast compares the same complete update sequence, whereas the four-event
grids mix different sequences.

Scope remains narrow.  The structural extension was designed after the
four-event \textsc{dose-insensitive} result was known, so it is prospective
for its new executions but is not an independent pre-four-event confirmation.
It is not pooled with Stage~0 and does not establish that dose insensitivity
diagnoses writer quality.

%% file: sections/appendices/fresh_editor_confirmation.tex
\subsection{Fresh published-editor confirmation}
\label{app:fresh-editor-confirmation}

This pre-outcome-frozen test uses 64 CounterFact families excluded from the
430-case previously accessed discovery pool by both case ID and normalized subject. Target
values may recur across panels; clean model reset for every family makes this
compatible with the frozen design.  The comparison holds GPT-2XL, the four
sequential singleton edits and their order, two held-out paraphrases, ten
neighborhood prompts, and the 10,000-draw paired family bootstrap fixed.  The
two operating points are sequential-singleton MEMIT-high (weight 6325) and
ROME-low (clamp factor 2), selected from discovery before confirmation outcome
access as the closest acquisition comparator.  They are not commensurate
strength labels.

\begin{table}[H]
\centering
\caption{Fresh 64-family confirmation.  $A$ is post-fourth-edit minus reset
target log probability; $L$ is terminal-minus-reset neighborhood success, so
$D=-L$ is locality damage. All brackets are pointwise 95\%
family-reweighting ranges; editor contrasts are formed within family before
reweighting.}
\small
\begin{tabular}{lccccc}
\toprule
Point & $A$ [95\% range] & $D$ & $L$ range & Install & Keep $\mid$ install \\
\midrule
MEMIT-high & $7.980$ $[7.272,8.651]$ & $3.750$ & $[-7.969,-0.781]$ & $0.508$ & $1.000$ \\
ROME-low & $7.522$ $[6.685,8.264]$ & $7.148$ & $[-11.523,-3.086]$ & $0.457$ & $0.966$ \\
\bottomrule
\end{tabular}
\label{tab:fresh-editor-confirmation}
\end{table}

Both points pass acquisition and retention but are inconclusive on the frozen
locality threshold, so neither full conjunction passes. The paired
MEMIT-minus-ROME acquisition contrast is $+0.459$ nats
(95\% range $[-0.093,+1.030]$).  The primary locality-success contrast is
$+3.398$ percentage points (95\% range $[-0.392,+7.070]$).  Because that range
crosses zero, the prespecified prediction that MEMIT would cause less
locality damage remains \textsc{inconclusive}.

Here the post-fourth-edit acquisition query is terminal, so continuous
retention duplicates acquisition rather than testing delay; installation and
retained-among-installs remain separate components.

Two read-only neighborhood diagnostics already emitted by the unchanged
runtime were secondary.  MEMIT changes the greedy next token on $6.5625\%$ of
queries versus $14.6875\%$ for ROME, a paired difference of $-8.125$
percentage points (95\% range $[-12.931,-3.750]$). Mean full-vocabulary
Kullback--Leibler (KL) divergence is
$0.1076$ versus $0.2136$, but the paired difference $-0.1060$ has 95\% range
$[-0.2707,+0.0180]$.  The greedy-change range lies below zero, but this
diagnostic does not replace
the inconclusive primary locality comparison or establish general safety.

%% file: sections/appendices/frozen_160_family_extension.tex
\subsection{Disjoint pre-outcome-frozen 160-family extension}
\label{app:new160-editor-confirmation}

\paragraph{Chronology and frozen panel.}
The earlier disjoint 64-family MEMIT-high--ROME-low comparison was
\textsc{inconclusive}: its paired locality-success contrast was $+3.3984$
percentage points with 95\% range $[-0.3916,+7.0703]$, and MEMIT did not clear its
own strict locality range.  That completed panel was used only as an
external pilot for sample size.  Its family-level paired standard deviation
was 15.1734 points. Assuming independent and identically distributed (i.i.d.)
families, a normal-approximation sizing heuristic gives approximately 157
families for 80\% two-sided power at the observed effect;
the design rounded upward to 160 for exact
eight-family shards and rank balance.  This estimate may be optimistic because
it is based on the pilot's observed effect.

The primary analysis uses only 160 new, disjoint CounterFact families and does
not pool the pilot.  Their 640 case IDs and normalized subjects have zero
overlap with all 686 previously accessed discovery and confirmation cases. The
fixed comparison is GPT-2XL at $d=4$, sequential-singleton MEMIT-high
(``mom2 update weight'' 6325) versus the discovery-selected,
acquisition-nearest ROME-low (clamp factor 2).  The protocol retains two
held-out paraphrases, ten neighborhood prompts, and four order replicates,
averaged within family.  The family is the reweighting unit; ranges use
10,000 paired family-bootstrap draws.

\begin{table}[H]
\centering
\caption{Disjoint new-160-family confirmation.  $A$ is query-time
acquisition in nats, $D=-L$ is terminal locality damage in percentage points,
and brackets are frozen 95\% empirical family-reweighting ranges.}
\label{tab:v83-new160-editor-confirmation}
\scriptsize
\begin{tabular}{lrrrrr}
\toprule
Editor & $A$ [95\% range] & $D$ ($L$ range) & Install & Retain/install & Full gate \\
\midrule
MEMIT-high & $8.3016\ [7.8805,8.7210]$ & $2.0781\ [-3.2031,-1.1250]$ & 0.4781 & 1.0000 & \textsc{pass} \\
ROME-low & $7.5166\ [7.0202,7.9752]$ & $4.6406\ [-6.4219,-3.0781]$ & 0.4172 & 0.9925 & \textsc{inconclusive}: locality \\
\bottomrule
\end{tabular}
\end{table}

The pre-outcome-frozen primary MEMIT-minus-ROME terminal locality-success
difference is $+2.5625$ percentage points with 95\% range
$[+1.0625,+4.2656]$.  It lies above zero, while MEMIT's own locality range
clears the frozen five-point gate and ROME's does not. MEMIT passes the
acquisition, retention, locality, and full gates; ROME passes acquisition and
retention, while its locality range crosses the threshold, making locality
and the full conjunction \textsc{inconclusive}. The prespecified confirmation
rule requires MEMIT's absolute pass and a paired locality range above zero;
ROME's inconclusive result remains specific to this rule and panel.
Acquisition is not forced into an
equivalence margin: the paired MEMIT-minus-ROME estimate is $+0.7850$ nats
with range $[+0.4416,+1.1441]$.

Because the dose-four query is terminal, continuous retention equals
acquisition and is not independent; installation and retained-among-installs
remain separate gates.

\paragraph{Secondary diagnostics.}
On preexisting read-only neighborhood outputs, MEMIT has lower terminal
full-vocabulary KL than ROME (paired difference $-0.094463$, 95\% range
$[-0.151609,-0.044748]$) and a lower greedy-next-token change fraction
($-0.111094$, $[-0.135156,-0.088750]$).  These diagnostics are secondary and
do not replace the pre-outcome-frozen locality-success comparison.

\paragraph{Execution checks and scope.}
Before reviewing outcomes, we found 152 completed runs and eight that failed
during model loading, before any update call. The eight failed runs were
rerun before outcome access with the same worker, model, strength, dose, and
estimator. All 160 planned runs and 1,280 rows were accounted for, and an
independent reanalysis reproduced the results.

On the new 160-family CounterFact panel, MEMIT-high met the prespecified
locality criterion and caused less locality damage than ROME-low. The
comparison uses one GPT-2XL checkpoint, dose four, and two editor
configurations selected from the discovery experiments. Sample size was
chosen using the earlier 64-family pilot, whose comparison remained
\textsc{inconclusive}.

%% file: sections/appendices/adaptive_rome_facet_sensitivity.tex
\subsection{Adaptive ROME-low eight-event facet}
\label{app:rome-eight-event}

\paragraph{Design and chronology.}
This secondary facet was designed after the four-event ROME grid and the
MEMIT-high eight-event result were known.  It therefore tests a newly frozen
ROME-low execution on a new structural extension; it is not an independent
pre-MEMIT comparison.  Sixteen disjoint CounterFact family pairs yield 32
focal orientations. For each focal orientation and order replicate, the four doses
share the exact eight-event sequence, giving 16 runs over the complete grid and
512 rows.
The inferential block is the disjoint family pair after averaging its two
orientations and four order replicates. Max-deviation ranges use 10,000
paired block-bootstrap draws, shared across doses.

\begin{table}[H]
\centering
\caption{Separately frozen ROME-low shared-eight-event facet. $A$, $R$, and $L$
denote acquisition, retention, and locality; $A$ is in nats and $D=-L$ is
terminal locality damage in percentage points. The locality endpoint and gate values are identical at
all four doses because the full ordered event sequence and terminal state are
identical. ``Install / retain'' reports installation and retention among
installs.}
\label{tab:v84-rome-eight-event}
\scriptsize
\begin{tabular}{crrrrl}
\toprule
Dose & $A$ & 95\% range for $A$ & $D$ & Install / retain & Gates \\
\midrule
1 & 6.37607 & $[5.45820,7.23267]$ & 6.71875 & 0.3125 / 0.8000 & A, R \textsc{pass}; L, full \textsc{inconclusive} \\
2 & 6.37283 & $[5.44759,7.24550]$ & 6.71875 & 0.3125 / 0.8000 & A, R \textsc{pass}; L, full \textsc{inconclusive} \\
4 & 6.28910 & $[5.38021,7.19332]$ & 6.71875 & 0.3125 / 0.8000 & A, R \textsc{pass}; L, full \textsc{inconclusive} \\
8 & 5.60322 & $[4.51207,6.67574]$ & 6.71875 & 0.3125 / 0.8000 & A, R \textsc{pass}; L, full \textsc{inconclusive} \\
\bottomrule
\end{tabular}
\end{table}

The frozen taxonomy is \textsc{inconclusive}. The observed acquisition range
is 0.77284 nats, but its max-deviation 95\% range
$[0.26121,1.28448]$ crosses the 0.5-nat practical threshold.  Dose 1 is the
descriptive peak, but its $d_1-d_2=+0.00324$-nat contrast has max-deviation
range $[-0.50840,+0.51488]$ and is unresolved.  The larger descriptive
fall to dose 8 thus cannot satisfy the prespecified
\textsc{dose-sensitive} conjunction, which requires both a range whose lower
endpoint exceeds the practical threshold and a winner contrast whose range
excludes zero.

Terminal-minus-reset retention is 5.60322 nats.  The locality-success range
is $[-12.65625,-1.48437]$ points at every dose.  Acquisition and retention
pass at every dose, whereas locality and the full conjunction are inconclusive
throughout because the range crosses the five-point threshold. For every
order replicate and focal orientation, we checked that the event order,
recorded reset/post-focal/terminal endpoints, retention, locality, recall,
and terminal parameters matched across doses. Only acquisition at the query
checkpoint varied. An independent reanalysis reproduced the results.

This facet is not pooled with either the four-event editor grid or the
MEMIT-high eight-event facet. It does not establish agreement between ROME and
MEMIT. Because one editor/strength's unresolved dose
surface and terminal gate status do not identify a general relationship
between schedule sensitivity and writer quality, it is not a writer-quality
diagnostic.

\subsection{Post-hoc relation-cluster sensitivity}
\label{app:v84-relation-sensitivity}

The frozen new-160 primary remains the paired family bootstrap: MEMIT-high
minus ROME-low terminal locality success is $+2.5625$ percentage points with a
95\% family-reweighting range $[+1.0625,+4.265625]$.  After outcomes were known, we audited whether
repeated CounterFact relations dominate that range.  The 160 focal
families contain 33 observed relations; 30 relations repeat, three are
singletons, 157/160 families lie in a repeated relation, and the largest
relation contains 12 families.

\begin{table}[H]
\centering
\caption{Post-hoc relation-aware sensitivity. Both ranges use a two-sided
Student-$t$ critical value with 32 degrees of freedom. CR1 denotes the
cluster-robust finite-sample correction. Neither range replaces the frozen
family-bootstrap primary; pp denotes percentage points.}
\label{tab:v84-relation-sensitivity}
\scriptsize
\begin{tabular}{lrrr}
\toprule
Sensitivity target & Estimate (pp) & SE (pp) & 95\% range (pp) \\
\midrule
Family-weighted point, relation-cluster CR1 & $+2.5625$ & 0.9353 & $[+0.6574,+4.4676]$ \\
Equal weight over 33 observed relation means & $+2.3621$ & 0.9405 & $[+0.4463,+4.2779]$ \\
\bottomrule
\end{tabular}
\end{table}

The CR1 check preserves the family-weighted point estimate and changes only
the uncertainty calculation.  The equal-relation check changes the
descriptive target to equal weight over the 33 observed relation means.  Both
post-hoc ranges lie above zero on this fixed GPT-2XL CounterFact panel, so
repeated focal relations do not explain away the observed advantage under
these two checks.  They do not create a pre-outcome-frozen relation-population
claim, establish a universal editor ranking, or generalize beyond the
selected model, benchmark, strengths, and endpoint.  The frozen paired-family
bootstrap remains the sole frozen decision summary.

%% file: sections/appendices/additional_editor_dose_study.tex
\subsection{Additional-editor dose breadth}
\label{app:additional-editor-breadth}

\paragraph{Frozen question and implementation check.}
On the inherited same-sequence eight-event GPT-2XL/CounterFact construction,
we prospectively predicted \textsc{safe+insensitive} for AlphaEdit
\citep{fang2024alphaedit} and \textsc{failing+sensitive} for EasyEdit's native
constrained fine-tuning baseline (FT-L). The first AlphaEdit implementation
reset a module-global cache on every
singleton and therefore did not execute faithful sequential AlphaEdit. Those
runs were stopped and excluded before outcome access. The corrected
implementation resets the cache only at sequence boundaries and carries it
across writes; it followed a pre-outcome implementation check but retained the
frozen predictions and estimator. FT-L was unaffected. In both valid cohorts,
each dose uses the same complete update sequence and the original acquisition,
retention, and locality criteria.

\begin{table}[H]
\centering
\scriptsize
\caption{Additional-editor breadth on the shared eight-event construction.
``Acquisition range'' is the max--min difference across doses; brackets are its
max-deviation-adjusted 95\% reweighting band. ``Safe'' denotes this CounterFact
neighborhood protocol only; pp denotes percentage points.}
\label{tab:v86-additional-editor-breadth}
\begin{tabularx}{\textwidth}{lYYYYY}
\toprule
Writer & Frozen prediction & Realized class & Acquisition range (nats) & Locality damage (pp) & Full-pass doses \\
\midrule
AlphaEdit & safe + insensitive & safe + insensitive & $0.00122$ $[0,0.01301]$ & $1.5625$ & $4/4$ \\
FT-L & failing + sensitive & failing + inconclusive & $0.61832$ $[0.08915,1.14749]$ & $12.8125$ & $0/4$ \\
\bottomrule
\end{tabularx}
\end{table}

AlphaEdit's dose-specific acquisition values at $d=1,2,4,8$ are
$7.719544$, $7.719592$, $7.719317$, and $7.718369$ nats.  Its joint-install
rate is $0.5625$, retained-per-install is $1.0$, and every dose clears the full
gate.  FT-L's corresponding acquisition values are $0.276407$, $0.246447$,
$0.133071$, and $-0.341910$ nats.  It has zero joint installs and no full-pass
dose.  FT-L's max-deviation range straddles the frozen 0.5-nat margin,
so its categorical sensitivity decision is \textsc{inconclusive}; independently,
zero joint installations and zero full-pass doses make its frozen
behavioral conjunction fail.

AlphaEdit met the behavioral criteria and was insensitive to dose, as
predicted. FT-L failed the behavioral criteria, while its dose sensitivity
remained inconclusive. The joint prediction was therefore not confirmed:
none of the tested published editors conclusively combined behavioral
failure with dose sensitivity. These classifications concern the tested
configurations and the declared CounterFact neighborhood protocol.

Across the shared-eight-event MEMIT, ROME, AlphaEdit, and FT-L facets, the
dose-eight query is terminal: continuous retention there equals acquisition and is
zero-lag, whereas doses one, two, and four include intervening edits.

%% file: sections/appendices/wikirecent_prospective_study.tex
\subsection{Prospective WikiRecent evaluation}
\label{app:wikirecent-prospective}

The panel, policies, tracing rule, estimands, bootstrap draws, decision rules,
and transport criterion were fixed before the preflight runs.
It is a knowledge-insertion study: each focal item has a new target, canonical
update prompt, one source-provided rephrase, and two relation-specificity
probes, but no old focal target. It is therefore not pooled with the three
replacement-task endpoint studies above.

\paragraph{Frozen design and resolution.}
The deterministic panel contains 64 families of four distinct edits. Four
cyclic Latin orders place each focal edit at every rank. MEMIT-high
($\texttt{mom2\_update\_weight}=6325$) and ROME-low
($\texttt{clamp\_norm\_factor}=2$) each execute four singleton GPT-2XL outer
calls, yielding $64\times4\times2=512$ policy rows and 256 paired
family--order units. Both held-out preflight examples passed before evaluation
outcomes were opened. For every executed row, the recorded outer calls matched
the precomputed score-ancestral sequence, giving $s_Z=1$ for the planned
MEMIT-minus-ROME contrast on canonical and rephrase scores.

Before outcomes, the allocator used only KnowEdit WikiRecent and the
GPT-2XL tokenizer, with no model scores or editor outputs. Schema, probe, and token constraints
preceded exclusion of prior-panel overlap in normalized subject, update prompt,
new target, or source record. It then sorted and greedily selected the first 256
rows with globally unique subjects, target strings and token sequences, and all
update, rephrase, and locality prompts. A four-item preflight panel was disjoint
from the evaluation panel under the same identifiers. The policies were
inherited from earlier GPT-2XL work and fixed before WikiRecent outcomes.
This study therefore tests the workflow using previously selected policies.

\begin{table}[H]
\centering
\caption{\textbf{Prospectively frozen effect and transport record.} Entries are
mean dose-four-minus-reset per-target-token log probabilities in nats. Brackets are
95\% empirical family-reweighting ranges from the fixed 64-family panel, not
population confidence intervals.}
\label{tab:wikirecent-effect}
\footnotesize
\begin{tabularx}{\textwidth}{p{0.20\textwidth}rrYc}
\toprule
Context & MEMIT-high & ROME-low & Mean paired difference & Effect status \\
\midrule
Canonical prompt & 4.1261 & 3.9119 & $+0.2142\ [+0.1079,+0.3369]$ & \textsc{pass} \\
One rephrase & 2.9104 & 2.1925 & $+0.7178\ [+0.5530,+0.8903]$ & \textsc{pass} \\
\bottomrule
\end{tabularx}
\end{table}

The pre-outcome-frozen transport predicate preserves only the resolved mean
direction under identity policy correspondence. Both ranges lie above zero,
so it passes. Descriptive post-hoc heterogeneity checks show that on canonical
prompts 35/64 family means and 115/256 order-pair contrasts are positive (141
are negative), the family median is $+0.0057$, and the focal-rank-four mean is
$-0.0158$ nats. On rephrases 58/64 family means are positive, but only 33/64
are positive in both contexts. This is aggregate-direction transport, not
family-, prompt-, order-, or rank-wise dominance.

\paragraph{Absolute gates.}
Table~\ref{tab:wikirecent-gates} keeps the relative effect separate from each
policy's acquisition, installation, retention-among-installs, and shared
locality records. Every non-locality component passes its pre-outcome-frozen rule.
ROME's rephrase installation point is 20.70\%, just above the 20\% threshold;
its family-reweighting range $[11.72\%,30.47\%]$ crosses that threshold because
the frozen installation gate is point-estimate based.

\begin{table}[H]
\centering
\caption{\textbf{Absolute feasibility fails despite transported mean
acquisition.} Acquisition is in nats; installation and retained-among-installs
are proportions; locality is terminal-minus-reset relation-specificity success
in percentage points. All entries are prospective.}
\label{tab:wikirecent-gates}
\scriptsize
\begin{tabularx}{\textwidth}{p{0.13\textwidth}p{0.12\textwidth}rrrrY}
\toprule
Context & Policy & Acquisition & Install & Retain/inst. & Locality & Full \\
\midrule
Canonical & MEMIT & 4.1261 & .8867 & .9912 & $-43.36\ [-52.73,-33.98]$ & \textsc{fail} \\
Canonical & ROME & 3.9119 & .8047 & .9563 & $-41.41\ [-50.78,-32.23]$ & \textsc{fail} \\
Rephrase & MEMIT & 2.9104 & .3828 & 1.0000 & $-43.36\ [-52.73,-33.98]$ & \textsc{fail} \\
Rephrase & ROME & 2.1925 & .2070 & .9811 & $-41.41\ [-50.78,-32.23]$ & \textsc{fail} \\
\bottomrule
\end{tabularx}
\end{table}

Locality here is deliberately narrow. Each of two source-supplied
relation-specificity probes contributes an indicator of whether the source-correct
answer has lower token-average negative log-likelihood (NLL);
their mean is the row score in both prompt contexts. The large negative changes
cross the pre-outcome-frozen $-5$-point
boundary for both policies. Consequently all four policy--context feasibility
conjunctions, the designated MEMIT target-feasibility claim, and
$\mathsf{EvidenceComplete}$ are \textsc{fail} even though mean direction
transport is \textsc{pass}.

On the same two probes and 256 family--order rows per editor, post-hoc terminal
full-vocabulary Kullback--Leibler (KL) divergence is 2.2944/2.4154 and greedy
next-token change is 73.63\%/77.73\% for MEMIT/ROME. Neither metric was a
frozen gate; because both reuse the same probes, they add no probe breadth and
support no population claim.

Continuous acquisition and terminal-minus-reset retention are numerically the
same dose-four checkpoint and provide only one continuous confirmation. The
prefix-retention summary also contains a zero-lag stratum when the focal edit
is fourth. Restricting descriptively to focal ranks 1--3, canonical retention
is 169/171 for MEMIT and 140/149 for ROME; rephrase retention is 73/73 and
37/38. The rank-four zero-lag counts for MEMIT and ROME are 56/56 and 57/57 on
canonical prompts and 25/25 and 15/15 on rephrases. Costs are recorded, but no common budget or scalarization
was declared and no cost gate is active.

\paragraph{Execution checks and reproducibility.}
The protocol and source manifest were fixed before both preflight runs.
Both preflight examples passed, and their results were collected before an
error occurred while publishing the result files. The publication path was
corrected using the existing files; neither preflight run was repeated or
recollected. A metadata timeout initially prevented one of the 64 evaluation
jobs from being submitted. We verified that the job had not been submitted,
then submitted its first attempt before accessing any evaluation outputs.
All 64 jobs completed successfully without restarts, and their outputs were
collected once.

All 15 checks of the prespecified protocol passed. An independent analysis
of the recorded rows reproduced the final results. The execution records
account for the submission correction and completion of all planned jobs.
The reproducibility bundle omits the remote scheduler records, so it cannot
independently reconstruct the submission and completion chronology.

%% file: sections/appendices/gpt2xl_endpoint_transport.tex
\subsection{GPT-2XL endpoint-native transport}
\label{app:endpoint-transport}

Three studies separately froze a fixed GPT-2XL operating point and tested it
with endpoint-native estimands. They are three fixed settings, not exchangeable
replicates: CounterFact uses two unseen rephrases and ten neighborhood prompts;
KnowEdit-zsRE uses one rephrase and two relation-specificity prompts; and
KnowEdit WikiData-CounterFact uses one rephrase and the first two canonical
source-ordered relation probes. Dataset and locality instrument therefore vary
jointly. All empirical reweighting ranges are computed separately by endpoint; no pooled
interaction is estimated, and the contrasts cannot identify a causal dataset
or instrument effect.

\paragraph{Frozen operating point.}
All three studies use the public Hugging Face GPT-2XL model, MEMIT-high
(\texttt{mom2\_\allowbreak update\_\allowbreak weight=\allowbreak6325}), ROME-low
(\texttt{clamp\_\allowbreak norm\_\allowbreak factor=\allowbreak2}), four sequential singleton edits, and
query/terminal dose $d=4$. The inferential family counts are 160 for
CounterFact, 160 for zsRE, and 128 for WikiData-CounterFact.

All runs use EasyEdit, with
\texttt{subject\_last}, value-loss layer 47, value learning rate 0.5,
configuration field \texttt{v\_num\_grad\_steps=20}, value weight decay 0.5,
KL factor 0.0625,
\texttt{transformer.h.\{\}.attn}, \texttt{transformer.h.\{\}.mlp}, and
\texttt{transformer.h.\{\}.mlp.c\_proj} modules,
\texttt{transformer.h.\{\}} layer modules, \texttt{transformer.ln\_f},
\texttt{transformer.wte}, Wikipedia moment statistics with 100,000 samples in
float32 from the stored statistics, device 0, and model parallelism
disabled. MEMIT-high edits layers 13--17 with all-layer selection, moment
adjustment enabled, moment-update weight 6325, and clamp factor 0.75. ROME-low
edits layer 17 with moment adjustment disabled, clamp factor 2, fp16 disabled,
and context-template length/count pairs $(5,10)$ and $(10,10)$.

\paragraph{Estimands, reweighting ranges, and gates.}
At reset and terminal, each relation probe is paired with a canonical source
answer and the focal edited target. We prepend one space to each continuation,
tokenize it without special tokens, sum its autoregressive token negative
log-likelihood, and divide by the number of continuation tokens. A probe
succeeds exactly when the source-correct answer has lower token-average
negative log-likelihood than the edited target. The compiler deterministically
keeps the first nonempty source-answer string (or the first string in the first
list); it neither maximizes nor marginalizes over aliases. Probe indicators are
averaged within row, then the four Latin-order replicates are averaged
within family at reset and terminal. Locality is
$L=100(\text{terminal success}-\text{reset success})$.

Each endpoint uses a paired family bootstrap: the same resampled family indices
are used for both editors, with 10,000 draws and linearly interpolated 2.5th and
97.5th percentiles.
The acquisition gate requires
its range's lower endpoint to exceed zero. Retention requires its range's
lower endpoint to exceed zero, joint install $\geq20\%$, and retained per
install $\geq50\%$; locality requires its point estimate to be at least $-5$
points and its range's lower endpoint to exceed $-5$.

The per-editor 95\% inputs to those absolute gates are as follows. On
CounterFact, MEMIT has acquisition/retention ranges
$[7.880546,8.721021]$, locality $[-3.203125,-1.125000]$, joint install
$[0.401563,0.554688]$, and retained per install $[1,1]$; ROME has
$[7.020214,7.975242]$, $[-6.421875,-3.078125]$,
$[0.346875,0.487500]$, and $[0.975709,1]$. On zsRE, the corresponding MEMIT
ranges are $[4.121185,4.816444]$, $[-37.343750,-25.390625]$,
$[0.509375,0.657813]$, and $[0.991279,1]$; ROME's are
$[3.555082,4.225588]$, $[-29.140625,-18.671875]$,
$[0.351523,0.485938]$, and $[0.963454,0.996032]$. On
WikiData-CounterFact, MEMIT has $[3.632744,4.373639]$,
$[-75.878906,-63.183594]$, $[0.072266,0.183594]$, and $[0.947368,1]$;
ROME has $[2.376302,3.042119]$, $[-60.839844,-46.579590]$,
$[0.031250,0.101563]$, and $[0.884615,1]$.

\begin{table}[H]
\centering
\caption{Fixed-strength GPT-2XL results under separately frozen
endpoint-native instruments. Acquisition $A$ is in nats and damage $D$ in
percentage points. The final column is the within-endpoint MEMIT-minus-ROME
locality-success contrast with a paired-family 95\% empirical reweighting range.}
\label{tab:v88-endpoint-transport}
\small
\begin{tabular}{llrrl}
\toprule
Endpoint & Editor & $A$ & $D$ & MEMIT--ROME locality success \\
\midrule
CounterFact & MEMIT-high & 8.3016 & 2.0781 & $+2.5625$ $[+1.0625,+4.2656]$ \\
CounterFact & ROME-low   & 7.5166 & 4.6406 & --- \\
KnowEdit-zsRE & MEMIT-high & 4.457329 & 31.328125 & $-7.5000$ $[-11.015625,-4.218750]$ \\
KnowEdit-zsRE & ROME-low   & 3.880083 & 23.828125 & --- \\
KnowEdit-WikiData-CF & MEMIT-high & 3.997292 & 69.726562 & $-15.9180$ $[-20.9961,-10.9375]$ \\
KnowEdit-WikiData-CF & ROME-low & 2.704324 & 53.808594 & --- \\
\bottomrule
\end{tabular}
\end{table}

\begin{sloppypar}
\paragraph{Results.}
MEMIT's acquisition advantage is positive in all three settings:
$+0.7850266$ nats $[+0.4416111,+1.1441018]$ on CounterFact,
$+0.5772456$ $[+0.4301146,+0.7341698]$ on zsRE, and
$+1.2929676$ $[+1.0650381,+1.5207201]$ on WikiData-CounterFact. The
locality-success contrasts instead are $+2.5625$, $-7.5000$, and $-15.9180$
points, respectively. Thus the locality ranking reverses on two endpoints,
whereas the acquisition ranking does not.
\end{sloppypar}

Before preflight or outcome access, the WikiData-CounterFact study predicted
a MEMIT locality advantage. The observed comparison favored ROME, as did
the zsRE comparison. Both results use the prespecified estimators.

On zsRE, both editors fail the absolute locality gate and therefore the
behavioral acquisition/retention/locality conjunction. MEMIT
installs 58.4375\% of focal cases and retains 99.7326\% per install; ROME
installs 41.7188\% and retains 98.1273\% per install. On
WikiData-CounterFact, both acquisition gates pass, but both retention gates fail
because joint install is 12.50\% for MEMIT and 6.25\% for ROME, below the frozen
20\% floor. Retained per install is 98.4375\% and 96.8750\%, respectively, but
cannot rescue that failed conjunct. Both the locality gate and the behavioral
acquisition/retention/locality conjunction fail. The
relative ROME advantage therefore does not make ROME safe. In all three
panels, the dose-four query is terminal, so continuous acquisition and retention
are numerically identical by construction rather than independent confirmations.

\paragraph{Execution checks.}
Across the three cohorts, retrospective reconstruction from the source and
execution records, excluding outcome values, matched all 448 planned
comparisons. Comparing complete request objects accounted for 1,792
family--order units, 3,584 ROME/MEMIT score-ancestral traces, 14,336
score-ancestral singleton calls, and another 14,336 post-reset calls used
for terminal validation. Completion checks accounted for every planned run.
In CounterFact, eight runs that failed during model loading were replaced
before outcome access using the same worker and scientific configuration
(Appendix~\ref{app:new160-editor-confirmation}); the KnowEdit cohorts
required no replacements. The reconstruction covers declared outer calls.
It does not provide an observed log for each call, per-event parameter
records, random-number-generator snapshots, or observed EasyEdit
optimization and solve steps. The configured step count above is an
implementation setting rather than an observed count of internal steps.
The reproducibility bundle supplies aggregate execution summaries but omits
private scheduler receipts, so these receipts cannot be independently
inspected through the bundle.

All 160/160 zsRE and 128/128 WikiData-CounterFact planned runs completed
without retry or replacement, producing 1,280/1,280 and 1,024/1,024 expected
editor--family--order rows. An independent reanalysis reproduced the estimates,
ranges, and observed locality reversals. The evaluated zsRE file contains
1,301 rows although its source card reports 1,230; the evaluated
WikiData-CounterFact file contains 839 although its card reports 885. Each
dataset file passed schema validation and was used in full.

For WikiData-CounterFact, the outcome-blind eligible pool contained 589
globally disjoint rows. The design froze 128 families (512 update records), the
largest launch-compatible power of two, without outcome-driven resizing.
Selected families have zero normalized subject, update-prompt, new-target, or
source-record overlap with the prior panels; 69 old-target strings repeat.
The evidence manifest links the source files, completion checks, collected
rows, and analyses.
The three positive acquisition contrasts and
$({+}2.5625,{-}7.5000,{-}15.9180)$
locality pattern are evidence from three fixed bundles, not a prevalence
estimate or a causal endpoint effect.

%% file: sections/appendices/llama_recurrence_test.tex
\subsection{Prospective Llama-3.2-3B recurrence test}
\label{app:llama-transport}

\paragraph{Prespecified question and model configuration.}
Before preflight or outcomes, we froze the question of whether GPT-2XL's
MEMIT-minus-ROME locality signature $({+},{-},{-})$ would recur, together with
the three panels, primary binary-locality estimator, paired family bootstrap,
prediction, and interpretation branches. The study uses
\texttt{unsloth/Llama-3.2-3B}, ROME's native
\texttt{clamp\_norm\_factor=4}, and MEMIT
\texttt{mom2\_update\_weight=15000}, without endpoint-specific strength
selection. Model and native configuration therefore change jointly; this is
not an isolated model-effect test.

\paragraph{Execution checks.}
All 448 predeclared runs completed on their first attempt, without retry or
failure, yielding 3,584 rows. Before collecting outcomes, we checked that
all 448 planned runs were complete and that their outputs corresponded to
the specified source and plan. An independent reanalysis reproduced the
results, and all 15 protocol checks passed.

\paragraph{Later outer-call reconstruction.}
A post-outcome structural projection that excludes outcome fields validates
3,584/3,584 declared atomic outer-call rows. At the dose-four/terminal score
cut, all 1,792 paired MEMIT--ROME traces separate ($s_Z=1$); reset is aliased
and the balanced pre-focal mixture has $s_Z=3/4$. EasyEdit internals and a
cross-bundle correspondence record remain unavailable. At $s_Z=1$ the
bounded-effect certificate spans the full outcome range, so this rules out only
declared outer-call aliasing. This diagnostic does not establish the full
correspondence required for formal transport between these model configurations.

\begin{table}[H]
\centering
\caption{\textbf{Frozen Llama-3.2-3B endpoint results.} Editor-level signed
locality (M: MEMIT; R: ROME) and paired M-minus-R contrasts. $L$ is
terminal-minus-reset success (points); $A$ is acquisition (nats). Brackets are
pointwise 95\% empirical family-reweighting ranges (10,000 paired draws);
ordering uses paired contrasts.}
\label{tab:llama-transport}
\scriptsize
\begingroup
\renewcommand{\arraystretch}{0.82}
\resizebox{\textwidth}{!}{%
\begin{tabular}{lrlll}
\toprule
Endpoint & Families & Editor-level $L$ [95\% range]
  & Paired $\Delta L$ [95\% range] & Paired $\Delta A$ [95\% range] \\
\midrule
CounterFact & 160
  & \begin{tabular}[c]{@{}l@{}}M: $-7.9531\ [-10.7188,-5.4688]$\\R: $-29.1719\ [-33.4688,-25.0313]$\end{tabular}
  & $+21.2188\ [+17.8438,+24.6094]$
  & $-0.3017\ [-0.5264,-0.1076]$ \\
KnowEdit-zsRE & 160
  & \begin{tabular}[c]{@{}l@{}}M: $-43.0469\ [-49.2969,-36.8750]$\\R: $-47.5000\ [-53.7500,-41.3281]$\end{tabular}
  & $+4.4531\ [+1.1719,+7.8125]$
  & $-0.2116\ [-0.3173,-0.1046]$ \\
KnowEdit WikiData-CounterFact & 128
  & \begin{tabular}[c]{@{}l@{}}M: $-70.5078\ [-75.7813,-65.1367]$\\R: $-70.5078\ [-75.4883,-65.4297]$\end{tabular}
  & $0.0000\ [-3.8086,+3.9063]$
  & $-0.2247\ [-0.4566,-0.0052]$ \\
\bottomrule
\end{tabular}}
\endgroup
\end{table}

\paragraph{Frozen decision.}
The predicted locality signature was $({+},{-},{-})$ and the realized one is
$({+},{+},{?})$. WikiData is unresolved, so the primary pattern test is
\textsc{inconclusive}. Excluding whole families whose new-target
first subtoken appears in their update prompt leaves 160/152/124
CounterFact/zsRE/WikiData-CounterFact families and gives the same signature.
All six method--endpoint points pass acquisition and adjacent
terminal retention but fail locality and the full conjunction; retention here
duplicates acquisition. Thus positive zsRE defeats exact recurrence, while
unresolved WikiData retains the \textsc{inconclusive} pattern call. Joint model
and native-configuration changes preclude architecture-specific attribution.

%% file: sections/appendices/end_to_end_and_multiplicity.tex
\setlength{\intextsep}{8pt plus 2pt minus 2pt}
\setlength{\abovecaptionskip}{6pt}
\renewcommand{\arraystretch}{1.0}
\small

\subsection{Combined results for the GPT-2XL comparison}
\label{app:end-to-end-record}

Table~\ref{tab:v93-complete-record} combines the existing results for the fixed
GPT-2XL comparison, with each endpoint's timing status reported separately.
All required evidence fields are populated, although
$\mathsf{EvidenceComplete}$ does not pass.

\begin{table}[H]
\centering
\caption{\textbf{Combined evidence for the fixed GPT-2XL comparison.} Resolution is conditional
on the outer-call adapter; cost has no declared gate; $H$ is effect-decision status.}
\label{tab:v93-complete-record}
\scriptsize
\begin{tabularx}{\textwidth}{>{\bfseries}p{0.13\textwidth}p{0.59\textwidth}Y}
\toprule
Record & Evidence & Status \\
\midrule
Resolution & Outer-call reconstruction gives $s_Z=1$ for 640 CounterFact, 640
zsRE, and 512 WikiData-CF family--order pairs. &
conditional \textsc{pass}; diagnostic \\
Effect & Source acquisition/locality contrasts are $+0.7850$
$[+0.4416,+1.1441]$ nats and $+2.5625$ $[+1.0625,+4.2656]$ points; endpoint
signs are $({+},{+},{+})$/$({+},{-},{-})$. & Acq. $H$: \textsc{undefined}
(no directional gate); locality $H$: \textsc{pass}/\textsc{fail}/\textsc{fail} \\
Feasibility & CounterFact: MEMIT passes and ROME is \textsc{inconclusive}; both
fail locality on zsRE and retention/locality on WikiData-CF. Costs are ungated. &
source mixed; targets \textsc{fail} \\
Transport & The correspondence fixes GPT-2XL, method identities, strengths,
four-call template, dose, and orientation while endpoint data/instruments
change. Locality reverses to $-7.5000$
$[-11.0156,-4.2188]$ and $-15.9180$ $[-20.9961,-10.9375]$. & \textsc{fail} \\
\bottomrule
\end{tabularx}
\end{table}

Timing and verification status are reported separately for each endpoint.
The retrospective reconstruction checks the worker, full requests, operator
identities, and four synchronous outer calls. It does not provide an
observed execution log for each call, EasyEdit's internal steps,
random-number-generator snapshots, parameters at each event, or the complete
model state.
Transport and the claim ladder fail; MEMIT fails target locality twice. Endpoint
and locality instrument co-vary, so no dataset-only causal effect is identified.

\subsection{Post-hoc across-endpoint sensitivity for the Llama signs}
\label{app:llama-multiplicity}

The frozen primary uses separate pointwise 95\% paired-family ranges. Post-hoc
Bonferroni ranges use endpoint-specific reweighting percentiles 0.8333 and
99.1667 (10,000 draws), without pooling. They test
simultaneous sign stability, not population coverage.

\begin{table}[H]
\centering
\caption{\textbf{Bonferroni sensitivity.} MEMIT-minus-ROME 98.333\% marginal
ranges span three endpoints; no population coverage is claimed.}
\label{tab:v93-llama-bonferroni}
\scriptsize
\begin{tabular}{lrr}
\toprule
Endpoint & Locality success (points) & Acquisition (nats) \\
\midrule
CounterFact & $+21.2188$ $[+17.2031,+25.4480]$ & $-0.3017$ $[-0.5827,-0.0730]$ \\
KnowEdit-zsRE & $+4.4531$ $[+0.3906,+8.5156]$ & $-0.2116$ $[-0.3405,-0.0808]$ \\
WikiData-CounterFact & $0.0000$ $[-4.6875,+4.8828]$ & $-0.2247$ $[-0.5071,+0.0441]$ \\
\bottomrule
\end{tabular}
\end{table}

Adjusted locality remains $({+},{+},{?})$: WikiData is unresolved and positive
zsRE defeats frozen $({+},{-},{-})$ recurrence. Acquisition becomes
$({-},{-},{?})$, so $({-},{-},{-})$ is only pointwise. The endpointwise rules
were prespecified; Bonferroni is a post-hoc simultaneous-stability test.

\subsection{Post-hoc cluster-pairs sensitivity}
\label{app:relation-cluster-sensitivity}

We resample paired family differences by relation cluster: 33 native CounterFact
IDs and 105 zsRE/27 WikiData-CounterFact exact subject-redacted templates.
KnowEdit lacks semantic IDs, so templates may split paraphrases.

\begin{table}[H]
\centering
\caption{\textbf{Post-hoc cluster-pairs sensitivity.} MEMIT-minus-ROME points
have Bonferroni-indexed 98.33\% ranges within each model--metric family;
CounterFact uses native IDs and KnowEdit exact templates.}
\label{tab:v93-cluster-sensitivity}
\scriptsize
\begin{tabular}{llrr}
\toprule
Bundle & Endpoint & Acquisition (nats) & Locality (points) \\
\midrule
GPT-2XL fixed & CounterFact & $+0.7850\ [+0.3779,+1.3489]$ & $+2.5625\ [+0.5719,+4.9598]$ \\
 & zsRE & $+0.5772\ [+0.4149,+0.7682]$ & $-7.5000\ [-12.0295,-3.2389]$ \\
 & WikiData-CF & $+1.2930\ [+1.0491,+1.5279]$ & $-15.9180\ [-23.4375,-10.1958]$ \\
Llama native & CounterFact & $-0.3017\ [-0.6196,-0.0818]$ & $+21.2188\ [+16.9091,+25.5673]$ \\
 & zsRE & $-0.2116\ [-0.3327,-0.0931]$ & $+4.4531\ [+0.4580,+8.7248]$ \\
 & WikiData-CF & $-0.2247\ [-0.4933,+0.1092]$ & $0.0000\ [-5.4314,+6.5476]$ \\
\bottomrule
\end{tabular}
\end{table}

All six GPT-2XL ranges retain their sign pair
$({+},{+},{+})/({+},{-},{-})$. Llama zsRE locality remains positive, WikiData
remains unresolved, and acquisition is $({-},{-},{?})$. This
checks exact relation-form dependence, not arbitrary semantic-relation or
model-population inference.

\normalsize

%% file: sections/appendices/formal_proofs.tex
\section{Complete proofs of the formal results}
\label{app:formal-proofs}

This section supplies the complete proofs referenced in the main text and by
the associated robustness discussions. The arguments below state all
measure-theoretic and linear-algebraic steps used by the named results.
Throughout, total variation uses the convention in
Definition~\ref{def:trace-resolution}.

\subsection{Deterministic history-resolution formula}
\label{app:proof-resolution-formula}

\begin{proof}[Derivation of Equation~\ref{eq:main-resolution}]
For row $r$, let $C_{r,1},\ldots,C_{r,K_r}$ be the history-equivalence
classes, so $(D_r^\top q)_k=\sum_{\pi\in C_{r,k}}q_\pi$. Under the positive
and negative contrast mixtures,
\[
P_{R,Z}^{q^\pm}(r,k)
=\frac{2w_r}{\lVert q\rVert_1}
  \sum_{\pi\in C_{r,k}}q_\pi^\pm .
\]
Since $q=q^+-q^-$, subtraction gives
\[
P_{R,Z}^{q^+}(r,k)-P_{R,Z}^{q^-}(r,k)
=\frac{2w_r}{\lVert q\rVert_1}(D_r^\top q)_k.
\]
The discrete total-variation identity therefore yields
\begin{align*}
\TV(P_{R,Z}^{q^+},P_{R,Z}^{q^-})
&=\frac12\sum_{r}\sum_{k=1}^{K_r}
  \left|\frac{2w_r}{\lVert q\rVert_1}(D_r^\top q)_k\right|\\
&=\frac{\sum_r w_r\lVert D_r^\top q\rVert_1}
        {\lVert q\rVert_1},
\end{align*}
which is Equation~\ref{eq:main-resolution}.
\end{proof}

\subsection{Count factorization}
\label{app:proof-count-sufficiency}

\begin{proof}[Proof of Lemma~\ref{lem:count-sufficiency}]
First suppose that $F(\Gamma)=Q(\nu(\Gamma))$ for every finite word. For any
$a,b\in\mathcal A$, the words $ab$ and $ba$ have the same count vector. Hence
\[
T_b\circ T_a=F(ab)=Q(\nu(ab))=Q(\nu(ba))=F(ba)=T_a\circ T_b.
\]
Thus the maps commute pairwise.

Conversely, suppose that the maps commute pairwise. Fix an ordering
$\mathcal A=\{a_1,\ldots,a_m\}$. For
$n=(n_1,\ldots,n_m)\in\mathbb N_0^m$, define
\[
Q(n):=T_{a_m}^{\,n_m}\circ\cdots\circ T_{a_1}^{\,n_1},
\]
where the zeroth power is the identity map. Any finite word $\Gamma$ can be
transformed into the canonical word
$a_1^{n_1}\cdots a_m^{n_m}$, with $n=\nu(\Gamma)$, by finitely many adjacent
swaps. Each adjacent swap preserves the composed map because the corresponding
two maps commute. Therefore
$F(\Gamma)=Q(\nu(\Gamma))$ for every finite word. This constructs the required
factorization and proves sufficiency.
\end{proof}

\begin{proof}[Restricted-catalogue factorization]
Let $\mathcal W$ be any restricted catalogue of words. If a factorization
$F|_{\mathcal W}=Q\circ\nu$ exists, then $F(\Gamma)=F(\Gamma')$ whenever
$\Gamma,\Gamma'\in\mathcal W$ have the same count vector. Conversely, if this
fiberwise equality holds, define $Q(n)$ for $n\in\nu(\mathcal W)$ to be
$F(\Gamma)$ for any $\Gamma\in\mathcal W$ with $\nu(\Gamma)=n$. The assumed
equality makes this definition independent of the chosen word. Extending $Q$
arbitrarily outside $\nu(\mathcal W)$ gives the desired factorization on
$\mathcal W$.
\end{proof}

The proof is map-valued: equality holds for every starting state in
$\mathcal X$. On a restricted word catalogue, factorization is equivalent only
to equality of the maps induced by admitted words sharing a count vector;
pairwise commutativity remains sufficient but is no longer necessary.

\subsection{History-resolution certificate}
\label{app:proof-trace-resolution-certificate}

We first record the elementary form of total-variation contraction used below.
If $P,Q$ are probability measures and $f$ is measurable with
$0\leq f\leq1$, the layer-cake representation gives
\[
\int f\,\mathrm d(P-Q)
=\int_0^1\{P(f>t)-Q(f>t)\}\,\mathrm dt,
\]
and hence
\begin{equation}
\left|\int f\,\mathrm d(P-Q)\right|\leq\TV(P,Q).
\label{eq:bounded-test-function-tv}
\end{equation}

\begin{proof}[Proof of Theorem~\ref{thm:trace-resolution-certificate}]
Write $X=(R,Z)$ and
$P_+=P_{R,Z}^{q^+}$, $P_-=P_{R,Z}^{q^-}$. Score sufficiency supplies a
single Markov kernel $L(\mathrm dy\mid x)$ for every policy in the contrast
support. Mixture linearity therefore gives the two joint laws
$P_+(\mathrm dx)L(\mathrm dy\mid x)$ and
$P_-(\mathrm dx)L(\mathrm dy\mid x)$, and their output marginals are
$Q_+=P_+L$ and $Q_-=P_-L$.

For every measurable output event $B$, the function
$f_B(x)=L(B\mid x)$ lies in $[0,1]$. Equation~\ref{eq:bounded-test-function-tv}
therefore implies
\[
|Q_+(B)-Q_-(B)|
=\left|\int f_B\,\mathrm d(P_+-P_-)\right|
\leq\TV(P_+,P_-).
\]
Taking the supremum over $B$ proves the Markov-contraction inequality
\[
\TV(Q_+,Q_-)\leq\TV(P_+,P_-)=s_Z(q).
\]

Because $Y_e\in[a,b]$, the rescaled function
$g(y)=(y-a)/(b-a)$ takes values in $[0,1]$. Applying
Equation~\ref{eq:bounded-test-function-tv} to $Q_+,Q_-$ and $g$ gives the
following; the additive constant $a$ cancels because both measures have total
mass one:
\begin{align*}
|\Delta_e^{\rm exec}(q)|
&=(b-a)\left|\int g\,\mathrm d(Q_+-Q_-)\right|\\
&\leq(b-a)\TV(Q_+,Q_-)\\
&\leq(b-a)s_Z(q).
\end{align*}

Finally, for a measurable summary $h$, every measurable set $C$ in its
codomain satisfies
\[
|h_\#P_+(C)-h_\#P_-(C)|
=|P_+(h^{-1}(C))-P_-(h^{-1}(C))|
\leq\TV(P_+,P_-).
\]
Taking the supremum over $C$ proves
$\TV(h_\#P_+,h_\#P_-)\leq s_Z(q)$.
\end{proof}

\begin{proof}[Equality condition for the summary contraction]
Let $\mu=P_+-P_-$ and let $A_+$ be a positive set in a Hahn decomposition of
$\mu$. Because $\mu$ has total mass zero,
$\mu(A_+)=\TV(P_+,P_-)$. If the sign partition is measurable through $h$, there
is a measurable set $C$ in the summary space such that
$h^{-1}(C)=A_+$ up to a $|\mu|$-null set. Consequently,
\[
h_\#P_+(C)-h_\#P_-(C)=\mu(A_+)=\TV(P_+,P_-).
\]
The reverse inequality is the data-processing bound already proved, so
equality holds. If $(R,Z)$ is recoverable almost surely from $h(R,Z)$ by a
measurable inverse, then every measurable sign set is measurable through $h$
up to a null set, which gives the stated sufficient special case.
\end{proof}

The common-initial-state, coupled-randomness, and complete-history conditions
in the main text are operational conditions intended to justify the common
kernel $L$. The proof itself requires that factorization; if an omitted path
makes the conditional outcome law depend on the policy after conditioning on
$(R,Z)$, the premise fails.

\subsection{Resolution floor}
\label{app:proof-resolution-floor}

\begin{proof}[Proof of Corollary~\ref{cor:resolution-floor}]
Theorem~\ref{thm:trace-resolution-certificate} and the asserted lower bound
give
\[
\delta\leq|\Delta_e^{\rm exec}(q)|
\leq(b-a)s_Z(q).
\]
Since $b-a>0$, division by $b-a$ yields
$s_Z(q)\geq\delta/(b-a)$. The contrapositive gives the stated pre-outcome
certificate.
\end{proof}

\subsection{Conditional-kernel mismatch}
\label{app:proof-kernel-mismatch}

\begin{proof}[Proof of Proposition~\ref{prop:kernel-mismatch}]
Write $P_+=P_X^{q^+}$ and $P_-=P_X^{q^-}$, and define
\[
Q_+=P_+K_+,
\qquad Q_-=P_-K_-,
\qquad Q_+^0=P_+L,
\qquad Q_-^0=P_-L.
\]
For every measurable output event $B$,
\begin{align*}
|Q_+(B)-Q_+^0(B)|
&=\left|\int\{K_+(B\mid x)-L(B\mid x)\}\,\mathrm dP_+(x)\right|\\
&\leq\int\TV(K_+(\cdot\mid x),L(\cdot\mid x))\,\mathrm dP_+(x)
=\epsilon_+.
\end{align*}
Taking the supremum over $B$ gives
$\TV(Q_+,Q_+^0)\leq\epsilon_+$. The same argument gives
$\TV(Q_-^0,Q_-)\leq\epsilon_-$. Markov contraction under the common kernel
$L$ gives
\[
\TV(Q_+^0,Q_-^0)\leq\TV(P_+,P_-)=s_Z(q).
\]
The triangle inequality now yields
\[
\TV(Q_+,Q_-)
\leq\epsilon_++s_Z(q)+\epsilon_-.
\]
Total variation between probability measures is at most one, proving the first
claimed bound after taking the minimum with one. Applying
Equation~\ref{eq:bounded-test-function-tv} to
$(Y_e-a)/(b-a)\in[0,1]$ proves the corresponding mean-score bound.
\end{proof}

\subsection{Schedule-class staircase}
\label{app:proof-mac-staircase}

\begin{proof}[Proof of Equation~\ref{eq:mac-staircase-v94} and the pairwise-resolution identity]
For $H=0$, every $a_p(H)$ is the empty sequence, so all $4!$ assignments have
one common history. For $H=1$, $a_1(H)=(1)$ while
$a_2(H)=a_4(H)=a_8(H)=(0)$. A history is therefore determined only by the
band receiving period one, giving four classes.

For $H\in\{2,3\}$, the sequences for periods one and two are distinct, while
$a_4(H)=a_8(H)$ is the all-zero sequence. Choosing the band receiving period
one and then the band receiving period two determines a history; swapping
periods four and eight does not change it. Hence there are $4\cdot3=12$
classes.

For every $H\geq4$, the entries at opportunities $1,2,4$ distinguish all four
periods: period one is active at opportunity one, period two is first active at
two, period four is first active at four, and period eight is inactive at all
three. Thus the four period sequences are distinct, so their assignments to
four labeled bands give $4!=24$ histories.

For a deterministic pair $q=\mathbf e_s-\mathbf e_{s'}$, the two joint laws put the same mass
$w_r$ on row $r$. If the two rowwise histories coincide, that row contributes
zero to their total variation. If they differ, the two distinct atoms each
contribute $w_r$ to the $\ell_1$ difference and hence contribute $w_r$ after
the factor one half. Summing over rows gives
\[
s_Z(q)=\sum_r w_r\ind\{z_{r,s,e}\neq z_{r,s',e}\}
=1-\sum_r w_r\ind\{z_{r,s,e}=z_{r,s',e}\}.
\]
\end{proof}

\subsection{Working-basis estimability}
\label{app:proof-working-basis-estimability}

\begin{proof}[Proof of Proposition~\ref{prop:working-basis-estimability}]
Each row of $D_r$ contains exactly one entry equal to 1, and each of its $K_r$ columns is
nonempty. Consequently, $D_r^\top D_r$ is diagonal with positive diagonal
entries and
$P_r=D_r(D_r^\top D_r)^\dagger D_r^\top$ is the orthogonal projector onto
$\operatorname{col}(D_r)$. Because
$\mathbf1=D_r\mathbf1_{K_r}$, it also satisfies
$P_r\mathbf1=\mathbf1$.

Define the observation operator
\[
\mathcal O v=(\sqrt{w_r}P_rv)_{r:w_r>0}.
\]
Two parameter pairs are observationally equivalent exactly when their
difference $(\delta\alpha,h)$ satisfies
$\mathcal O(\delta\alpha\mathbf1+Uh)=0$. Since every $P_r$ is symmetric and
idempotent,
\begin{align*}
\lVert\mathcal O(\delta\alpha\mathbf1+Uh)\rVert^2
&=\sum_r w_r\lVert P_r(\delta\alpha\mathbf1+Uh)\rVert^2\\
&=|\Pi|(\delta\alpha)^2
  +2\delta\alpha\sum_rw_r\mathbf1^\top P_rUh
  +h^\top U^\top\!\left(\sum_rw_rP_r\right)Uh\\
&=|\Pi|(\delta\alpha)^2+h^\top I_Uh.
\end{align*}
The cross term vanishes because
$\mathbf1^\top P_rU=(P_r\mathbf1)^\top U=\mathbf1^\top U=0$.
Both remaining terms are nonnegative. Thus an invisible perturbation must have
$\delta\alpha=0$ and $h^\top I_Uh=0$. Since $I_U$ is symmetric positive
semidefinite, the latter condition is equivalent to $h\in\ker(I_U)$.

The functional $a^\top\beta$ is structurally estimable exactly when
$a^\top h=0$ for every invisible $h$, that is, exactly when
$a\in(\ker I_U)^\perp$. For a symmetric matrix,
$(\ker I_U)^\perp=\operatorname{range}(I_U)$, which proves the claim.
\end{proof}

\subsection{Dimension of the band--period working basis}
\label{app:proof-permutation-dimension}

\begin{proof}[Proof of the nine-dimensional affine-span claim]
The difference of any two $4\times4$ permutation matrices has zero row sums
and zero column sums. The vector space
\[
\mathcal V=\{H\in\R^{4\times4}:H\mathbf1=0,
\ \mathbf1^\top H=0\}
\]
has dimension nine: its upper-left $3\times3$ block is free, the last entry in
each of the first three rows and columns is then forced, and the bottom-right
entry is forced consistently.

For $i,j\in\{1,2,3\}$, define
$B_{ij}=E_{ij}-E_{i4}-E_{4j}+E_{44}$. These nine matrices form a basis of
$\mathcal V$. To express $B_{ij}$ as a difference of permutation matrices,
choose any bijection between the two rows outside $\{i,4\}$ and the two
columns outside $\{j,4\}$. Let one permutation use the assignments
$i\mapsto j$ and $4\mapsto4$, and let the other use
$i\mapsto4$ and $4\mapsto j$, with the chosen remaining assignments shared.
Their matrix difference is exactly $B_{ij}$. Hence differences of permutation
matrices span $\mathcal V$. The affine hull of the 24 permutation matrices
therefore has dimension nine.
\end{proof}

\subsection{Off-support non-identification}
\label{app:proof-off-support}

\begin{proof}[Construction for Remark~\ref{rem:off-support-transport}]
Let the bounded outcome range contain distinct values $y_0$ and $y_1$, and fix
any response surface on the evaluated support $\mathcal H$. Extend it to two
bounded response surfaces $f_0$ and $f_1$ that agree with the fixed surface on
$\mathcal H$, agree with each other at every other point except
$x^\dagger$, and satisfy
$f_0(x^\dagger)=y_0$ and $f_1(x^\dagger)=y_1$. The two surfaces induce the
same complete record on $\mathcal H$ but different responses at
$x^\dagger$. The off-support response is therefore not identified without an
additional restriction or target evidence.

If a constant-sign hypothesis held on all mapped cells, every resolved cell
would have that same sign. Observing one resolved positive sign and one
resolved negative sign therefore contradicts the hypothesis. A cell whose
interval contains zero supplies neither sign and does not by itself establish
or refute constancy.

For the two-coordinate illustration in
Equation~\ref{eq:missing-crossed-cell}, the unobserved value $Y_{11}$ may be any
element of $[a,b]$. Subtracting the fixed quantity
$Y_{10}+Y_{01}-Y_{00}$ maps that interval exactly to the interval displayed in
the equation. Likewise, if the product in
Equation~\ref{eq:ranking-reversal} is negative, the two within-context utility
differences have opposite signs. Hence neither of the two constant policy
choices maximizes utility in both contexts within the stated two-policy set.
\end{proof}